\documentclass[runningheads]{llncs}

\usepackage{eccv}

\usepackage{eccvabbrv}

\usepackage{graphicx}
\usepackage{booktabs}
\usepackage{amsmath,amssymb}
\usepackage{subcaption}
\usepackage{enumitem}
\usepackage[protrusion=true,expansion=false]{microtype}  

\usepackage[accsupp]{axessibility}  

\usepackage{hyperref}

\usepackage{orcidlink}

\begin{document}

\title{DAPS++: Rethinking Diffusion Inverse Problems with Decoupled Posterior Annealing}

\titlerunning{DAPS++: Decoupled Posterior Annealing}

\author{Hao Chen\orcidlink{0009-0004-9430-2057} \and
Renzheng Zhang\orcidlink{0000-0002-0137-3533} \and
Scott S. Howard\orcidlink{0000-0003-3246-6799}}

\authorrunning{H.~Chen et al.}

\institute{
University of Notre Dame, Notre Dame, IN 46556, USA \\
\email{\{hchen27, rzhang4, showard\}@nd.edu}
}

\maketitle

\begin{abstract}
From a Bayesian perspective, score-based diffusion solves inverse problems through joint inference, embedding the likelihood with the prior to guide the sampling process. However, this formulation fails to explain its practical behavior: the prior offers limited guidance, while reconstruction is largely driven by the measurement-consistency term, leading to an inference process that is effectively decoupled from the diffusion dynamics. We show that the diffusion prior in these solvers functions primarily as a warm initializer that places estimates near the data manifold, while reconstruction is driven almost entirely by measurement consistency. Based on this observation, we introduce \textbf{DAPS++}, which fully decouples diffusion-based initialization from likelihood-driven refinement, allowing the likelihood term to guide inference more directly while maintaining numerical stability and providing insight into why unified diffusion trajectories remain effective in practice. By requiring fewer function evaluations (NFEs) and measurement-optimization steps, \textbf{DAPS++} achieves high computational efficiency and robust reconstruction performance across diverse image restoration tasks.
\end{abstract}

\section{Introduction}
\label{sec:intro}

Solving ill-posed inverse problems is fundamental in science and engineering, particularly in scientific imaging—such as astrophotography~\cite{porth2019event,lucy1994optimum} and biomedical imaging~\cite{song2021solving,fish1995blind,dey2006richardson,lustig2007sparse,jalal2021robust}—with broad applications in real-world systems. These problems aim to reconstruct the original signal $\mathbf{x}_0$ from observations $\mathbf{y}$ acquired under various restoration or measurement conditions, assuming a known forward model~\cite{tarantola2005inverse}. A common strategy is to formulate the inverse problem within a Bayesian framework~\cite{dashti2015bayesian}, where $p(\mathbf{x}_0|\mathbf{y}) \propto p(\mathbf{y}|\mathbf{x}_0)p(\mathbf{x}_0)$. Here, the likelihood $p(\mathbf{y}|\mathbf{x}_0)$ characterizes the measurement process, while the prior $p(\mathbf{x}_0)$ encodes structural assumptions about the underlying signal.

Classical Bayesian inverse problem formulations often rely on Tikhonov, total variation (TV), or wavelet-based priors to constrain the solution space and stabilize reconstruction~\cite{stuart2010inverse,gibbs2002choosing}. Modern generative models~\cite{ho2020denoising,song2020denoising,shah2018solving}, particularly score-based diffusion methods~\cite{song2019generative,song2021scorebased,song2020improved,karras2022elucidating}, provide far more expressive high-dimensional priors and have become widely adopted in Bayesian inverse problem frameworks~\cite{chung2022diffusion,zhang2025improving,feng2023score,cardoso2023monte}. These methods incorporate measurement information into the sampling dynamics by injecting an approximate likelihood gradient $\nabla_{\mathbf{x}_t}\log p(\mathbf{y}|\mathbf{x}_t)$ at each diffusion timestep. By solving stochastic differential equations (SDEs) or their deterministic ODE counterparts~\cite{lu2022dpm,lu2025dpm}, diffusion models transform an initial noise sample $\mathbf{x}_T$ toward the data manifold $\mathbf{x}_0$, conditioned on the observation $\mathbf{y}$.

More recently, decoupled formulations such as DCDP~\cite{li2024decoupled} and DAPS~\cite{zhang2025improving} explicitly separate the diffusion prior $p(\mathbf{x}_0)$ from the data-consistency term $p(\mathbf{y}|\mathbf{x}_0)$ at the algorithmic level. In these methods, diffusion updates and measurement updates alternate: each data-consistency step moves the estimate toward the observation, and the subsequent re-noising step restores the appropriate annealed noise level. As the noise level decreases, the resulting time-marginal distribution $\pi(\mathbf{x}_t|\mathbf{y})$ is assumed to approach the posterior $p(\mathbf{x}_0|\mathbf{y})$. Although implemented in a decoupled manner, the overall framework remains nominally Bayesian, since each marginal still depends on the prior gradient $\nabla_{\mathbf{x}_t}\log p_t(\mathbf{x}_t)$.

However, this expected trajectory rarely holds in practice. The gradient of data likelihood gradient often dominates the update dynamics, causing each intermediate distribution $\pi(\mathbf{x}_t|\mathbf{y})$ to act as an isolated approximation of the posterior rather than part of a coherent sequence of marginals. As a result, the path from $\mathbf{x}_T$ to $\mathbf{x}_0$ becomes a set of loosely connected estimates rather than a consistent posterior evolution. This reveals that a strictly Bayesian interpretation is insufficient for explaining the empirical behavior of diffusion-based inverse problem solvers. This reveals that the diffusion prior and the data-consistency update serve fundamentally different roles: the diffusion stage provides a structured initialization that constrains the feasible solution space to lie near the data manifold, while the likelihood gradient refines this initialization to satisfy the measurement. The two stages are thus naturally decoupled, corresponding to prior-driven initialization followed by likelihood-driven correction.

Building on this perspective, we introduce DAPS++. Leveraging the principles of Decoupled Annealing Posterior Sampling (DAPS), our method provides a more efficient and interpretable formulation of diffusion-based Bayesian inference using diffusion models that operate directly in pixel space. Empirically, DAPS++ delivers strong performance across a broad range of inverse problems, including both linear and nonlinear image restoration, while offering substantially faster sampling than existing approaches. Compared with DAPS and other diffusion-based inverse problem methods, DAPS++ achieves comparable or superior reconstruction quality while reducing sampling time and neural function evaluations (NFEs) by approximately 90\%on both the FFHQ validation set and the ImageNet test dataset.

\section{Related Work}
\label{sec:rw}

\subsection{Diffusion Models}
Score-based diffusion models~\cite{song2019generative,song2021scorebased,song2020improved,karras2022elucidating,ho2020denoising} learn the data distribution $p(\mathbf{x}_0)$ and its gradient, generating samples by progressively denoising Gaussian corrupted data. A sequence of isotropic Gaussian perturbations $\mathcal{N}(0,\sigma_t^2\mathbf{I})$ is applied to the data, inducing a family of noise-conditioned distributions $p(\mathbf{x};\sigma_t)$. The model is trained to estimate the gradient of the log-density as a \emph{score function} across time steps $t\in[0,T]$, or equivalently across decreasing noise levels $\sigma_t$ from $\sigma_T=\sigma{\max}$ to $\sigma_0=0$. When $\sigma_{\max}$ is sufficiently large, the terminal distribution approaches pure Gaussian noise, $\mathcal{N}(0,\sigma_{\max}^2\mathbf{I})$.

During sampling, the learned model approximates the true score of $p_t(\mathbf{x};\sigma_t)$ via  
$s_\theta(\mathbf{x},\sigma_t)\approx\nabla_{\mathbf{x}}\log p_t(\mathbf{x};\sigma_t)$.  
Starting from pure noise $\mathbf{x}_T\sim\mathcal{N}(0,\sigma_{\max}^2\mathbf{I})$, the generative process follows the reverse stochastic differential equation (SDE)~\cite{lu2022dpm,song2019generative}
\begin{equation}
    d\mathbf{x}_t
    = -2\,\dot{\sigma}_t\,\sigma_t\,\nabla_{\mathbf{x}_t}\log p_t(\mathbf{x}_t;\sigma_t)\,dt
      + \sqrt{2\,\dot{\sigma}_t\,\sigma_t}\, d\mathbf{w}_t ,
\label{eq:SDE_Diffusion}
\end{equation}
where $\dot{\sigma}_t$ is the derivative of the noise schedule, $d\mathbf{w}_t$ is a standard Wiener process, and the formulation follows the variance-exploding (VE) or EDM parameterization.

\subsection{Inverse Problems with Diffusion Models}
A general inverse problem seeks to recover the unknown signal $\mathbf{x}_0$ from noisy ill-posed measurements $\mathbf{y}$. Given a forward operator $\mathcal{A}$, the general measurement model is  
$\mathbf{y}=\mathcal{A}(\mathbf{x}_0)+\mathbf{n}$,  
where $\mathbf{n}$ is additive Gaussian noise $\mathcal{N}(0,\gamma^2\mathbf{I})$.  
Under a Bayesian formulation, the posterior satisfies  
$p(\mathbf{x}_0|\mathbf{y})\propto p(\mathbf{y}|\mathbf{x}_0)p(\mathbf{x}_0)$,  
with the likelihood term  
$p(\mathbf{y}|\mathbf{x}_0)=\mathcal{N}(\mathcal{A}(\mathbf{x}_0),\gamma^2\mathbf{I})$.

To incorporate the diffusion prior into inverse problems, a standard approach is to form a joint SDE by adding the data-consistency term to~\cref{eq:SDE_Diffusion}, giving
\begin{equation}
\label{eq:SDE:joint}
\begin{split}
d\mathbf{x}_t
&= -2\,\dot{\sigma}_t\,\sigma_t\,
    \nabla_{\mathbf{x}_t}\log p_t(\mathbf{x}_t;\sigma_t)\,dt \\
&\quad -2\,\dot{\sigma}_t\,\sigma_t\,
    \nabla_{\mathbf{x}_t}\log p_t(\mathbf{y}\mid\mathbf{x}_t;\sigma_t)\,dt \\
&\quad + \sqrt{2\,\dot{\sigma}_t\,\sigma_t}\, d\mathbf{w}_t
\end{split}
\end{equation}
which combines the diffusion prior drift with a measurement-consistency term.

In practice, measurement consistency is defined in terms of the clean likelihood $p(\mathbf{y}|\mathbf{x}_0)$; as a result, the noised likelihood $p(\mathbf{y}|\mathbf{x}_t;\sigma_t)$ is typically inaccessible or ill-defined. Several approaches approximate this quantity~\cite{song2024Resample,Choi_2021_ICCV} to embed measurement guidance directly into the diffusion process. Among these, Diffusion Posterior Sampling (DPS)~\cite{chung2022diffusion} employs Tweedie’s formula to estimate $p(\mathbf{y}|\mathbf{x}_t)\approx p(\mathbf{y}|\mathbf{x}_0=\mathbb{E}[\mathbf{x}_0|\mathbf{x}_t])$, achieving strong empirical performance. However, recent studies show that DPS behaves more like a maximum a posteriori (MAP) estimator than a true posterior sampler~\cite{xu2025rethinking,Yang2024Guidance}. This has prompted data-consistency frameworks with Plug-and-Play (PnP) priors~\cite{li2024decoupled,xu2024PnP,wu2024principled} that incorporate explicit likelihood optimization into diffusion-based solvers. These PnP-based methods use diffusion models as learned denoising priors within an explicit data-consistency loop, rather than depending on measurement-guided updates along the diffusion manifold. Nevertheless, approaches that minimize $\nabla_{\mathbf{x}_t}\log p_t(\mathbf{y}|\mathbf{x}_t;\sigma_t)$ often exhibit instability due to overfitting. Another line of work uses SVD-based projections and projection-driven sampling~\cite{wang2023DDNM,kawar2022DDRM,chung2022improving} to enforce measurement constraints, while recent decoupled frameworks such as DAPS~\cite{zhang2025improving} reformulate~\cref{eq:SDE:joint} into a two-stage structure that preserves the time-marginal distribution.

However, similar to DPS, such decoupled methods deviate from posterior sampling as defined in~\cref{eq:SDE:joint}, because they implicitly assume that the time-marginal distributions remain consistent with the true posterior given $\hat{\mathbf{x}}_0(\mathbf{x}_t)$, which does not hold in practice. In reality, as we show below from Sec.~\ref{sec:method:4}, the two stages operate as separate generation and optimization procedures rather than components of a unified posterior sampler, limiting both the theoretical validity and computational efficiency of the framework. This observation motivates reinterpreting diffusion-based inverse problem solvers primarily as methods for optimizing $p(\mathbf{y}|\mathbf{x}_0)$ within a diffusion-generated solution space, offering a more accurate perspective on diffusion-based inference.

\section{Method}
\label{sec:method}

\subsection{The Diffusion Prior as a Warm Initializer}

We begin by analyzing the interaction between the prior and likelihood gradients to clarify the functional role of the diffusion prior in inverse problem solving.

To quantify the directional relationship between the two terms, we examine the empirical inner product
\begin{equation}
A_t =
\Big\langle
\nabla_{\mathbf{x}_t}\log p_t(\mathbf{y}\mid\mathbf{x}_t;\sigma_t),\,
\nabla_{\mathbf{x}_t}\log p_t(\mathbf{x}_t;\sigma_t)
\Big\rangle ,
\label{eq:inner_prod}
\end{equation}
where the two factors denote the likelihood and prior score at noise level $\sigma_t$, respectively.
Under high-noise conditions, this inner product is empirically close to zero, indicating that the two gradients are effectively orthogonal.
This orthogonality helps explain why methods such as DPS can combine both terms without destructive interference (see~\cref{sec:method:4}).

To assess the relative strength of the two components, we define the gradient ratio
\begin{equation}
\kappa_t :=
\frac{
\left\|\nabla_{\mathbf{x}_t}\log p_t(\mathbf{y}\mid\mathbf{x}_t;\sigma_t)\right\|
}{
\left\|\nabla_{\mathbf{x}_t}\log p_t(\mathbf{x}_t;\sigma_t)\right\|
}.
\end{equation}
Under the Gaussian noise model
$p(\mathbf{y}|\mathbf{x}_0) =
\mathcal{N}(\mathcal{A}(\mathbf{x}_0), \gamma^2\mathbf{I})$,
the likelihood gradient at noise level $\sigma_t$ can be approximated as
\begin{equation}
\nabla_{\mathbf{x}_t} \log p_t(\mathbf{y}|\mathbf{x}_t; \sigma_t)
\simeq
-\frac{1}{\gamma^2}\,
\nabla_{\mathbf{x}_t}
\left\| \mathbf{y} - \mathcal{A}\big(\hat{\mathbf{x}}_0(\mathbf{x}_t)\big) \right\|^2,
\label{eq:likelihood_grad}
\end{equation}
where $\mathcal{A}(\cdot)$ is the measurement operator and
$\hat{\mathbf{x}}_0(\mathbf{x}_t)$ denotes the denoised estimate associated with state $\mathbf{x}_t$.

\begin{figure}[t]
  \centering
  \begin{minipage}[t]{0.54\linewidth}
    \centering
    \includegraphics[width=0.85\linewidth]{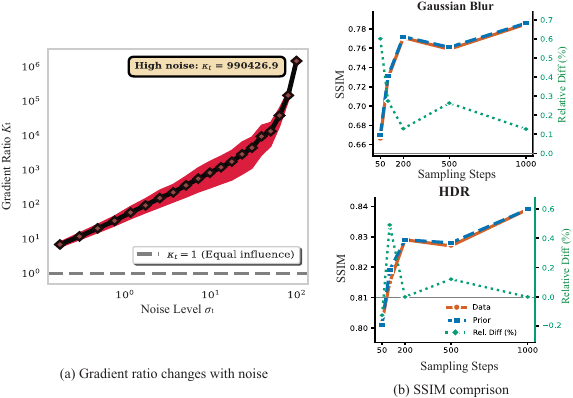}
    \caption{(a) Evolution of gradient ratio $\kappa_t$ vs.\ noise level during a Gaussian-blur iteration, showing data-consistency gradient dominates throughout. (b) Reconstruction quality of DAPS when driven by the \textbf{prior} (time-marginal) vs.\ the \textbf{data likelihood} term only across annealing steps, showing that data likelihood alone accounts for nearly all reconstruction quality.}
    \label{fig:gradient_ratio}
  \end{minipage}
  \hfill
  \begin{minipage}[t]{0.45 \linewidth}
    \centering
    \includegraphics[width=0.85\linewidth]{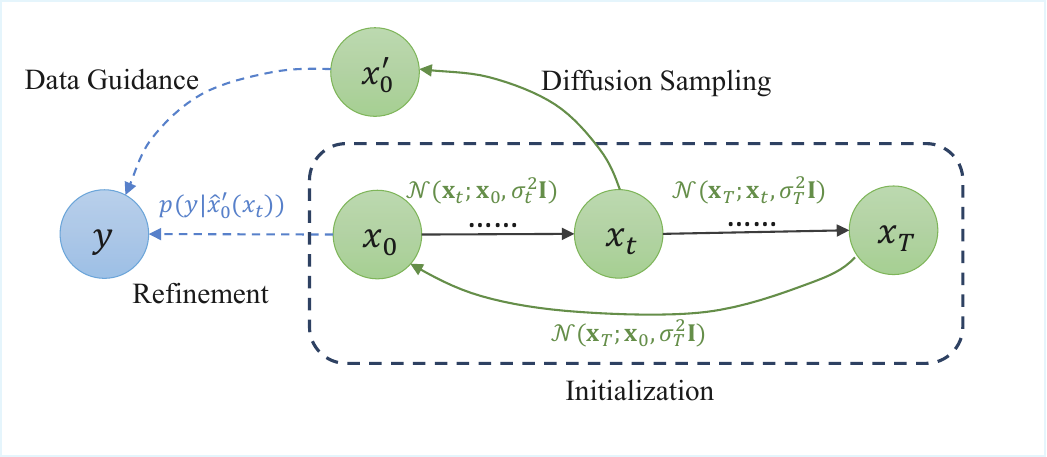}
    \caption{Diagram of the \textbf{DAPS++} framework. \textbf{Stage~1: Initialization} constructs the constrained optimization space $p(\mathbf{x}_0)$. \textbf{Stage~2: Refinement} optimizes under measurement guidance. The two stages are fully decoupled, with noise re-injected between steps.}
    \label{fig:dig}
  \end{minipage}
\end{figure}

As shown in~\cref{fig:gradient_ratio}(a), $\kappa_t \gg 1$ throughout the diffusion process.
A Lipschitz-bound analysis in the Appendix (Sec.~6.1) formalizes this: under standard regularity assumptions, $\kappa_t \gtrsim \frac{\sigma_{\min}^+(\mathcal{A})}{\gamma^2 C}\sigma_t|\mathbf{r}_t|$, which grows with the noise-to-measurement ratio $\sigma_t/\gamma$.
This does not imply that the diffusion prior is unnecessary; instead, it is essential for placing the initial estimate near the data manifold. The analysis shows that once initialization is achieved, the prior gradient contributes negligibly to subsequent updates: it is small in magnitude ($\kappa_t \gg 1$) and approximately orthogonal to the likelihood gradient ($A_t \approx 0$), meaning it neither accelerates nor redirects the measurement-driven correction. As confirmed in~\cref{fig:gradient_ratio}(b), removing the prior term from the time-marginal update and relying solely on the data-consistency term produces nearly identical reconstruction quality. This demonstrates that the continuous prior guidance used in existing methods can be effectively replaced by a one-time diffusion-based initialization.

\subsection{Complete Stage Separation: Initialization and Refinement}
\label{sec:method:sep}

The observation above suggests that decoupled noise-annealing approaches do not strictly follow the time-marginal distribution, especially when initialized from high-noise conditions.
Nevertheless, they produce strong reconstructions empirically because the diffusion model provides a high-quality initialization that the likelihood gradient can then refine.
This motivates a complete separation between the two stages, as illustrated in~\cref{fig:dig}.

\textbf{Stage 1: Diffusion Initialization.}
We start from $\mathbf{x}_T \sim \mathcal{N}(0, \sigma_{\max}^2\mathbf{I})$ and run
the reverse diffusion process without likelihood guidance.
A key design question is what form the initialization should take.
In the EDM noise schedule, the probability flow ODE is
\begin{equation}
d\mathbf{x} = -\dot{\sigma}\sigma\,\nabla_{\mathbf{x}}\log p(\mathbf{x};\sigma)\,dt,
\label{eq:pf_ode}
\end{equation}
and a single Euler step from $\mathbf{x}_t$ recovers exactly Tweedie's formula:
\begin{equation}
\mathbb{E}[\mathbf{x}_0 \mid \mathbf{x}_t] \;\approx\; \mathbf{x}_t \;+\; \sigma_t^2\, s_\theta(\mathbf{x}_t,\sigma_t),
\label{eq:Tweedie}
\end{equation}
where $s_\theta(\mathbf{x}_t,\sigma_t)$ denotes the learned score function.
As a first-order mean estimator, Tweedie's formula produces an initialization near the data manifold that is smooth and fast to compute.
Higher-order ODE solvers follow the diffusion trajectory more faithfully but do not improve Stage~2 convergence: what matters for refinement is proximity to the manifold, not precision of the sampling path, since the likelihood gradient handles the remaining correction. Figure~S1 in the Supplementary Material validates this by comparing initialization strategies across different sampling method settings, confirming that a single Tweedie step at an appropriate ${\sigma}$ provides a more effective starting point than ODE-based alternatives.

To balance efficiency and fidelity, DAPS++ introduces a noise threshold $\bar{\sigma}$: for $\sigma_t > \bar{\sigma}$, Tweedie denoising (\cref{eq:Tweedie}) provides fast initialization; for $\sigma_t \le \bar{\sigma}$, a single higher-order ODE step (RK4) preserves fine details at minimal additional cost.
The resulting $\hat{\mathbf{x}}_0$ captures the global structure of the signal and lies near the high-density region of $p(\mathbf{x}_0)$. Stage~2 requires only that $\hat{\mathbf{x}}_0$ is close to the data manifold; the specific initialization path is irrelevant, ensuring complete decoupling between the two stages.

\textbf{Stage 2: Likelihood-Driven Refinement.}
Given the initialization $\hat{\mathbf{x}}_0$, we perform MCMC sampling driven solely by the measurement likelihood.
Under the Gaussian measurement model, the likelihood gradient is globally Lipschitz and the log-density is strongly concave, giving the unadjusted Langevin algorithm (ULA) well-established convergence guarantees that depend on the specific type of regularization~\cite{10.1111/rssb.12183,AAP1238}.
Following ULA, the update rule is
\begin{equation}
\begin{aligned}
\mathbf{x}_0^{(j+1)}
&= \mathbf{x}_0^{(j)}
 + \eta\Big(
      \nabla_{\mathbf{x}_0^{(j)}}\log p(\mathbf{y}\mid\mathbf{x}_0^{(j)}) \\
&\qquad\qquad\ \
    + \nabla_{\mathbf{x}_0^{(j)}}\log p(\mathbf{x}_0^{(j)})
    \Big)
 + \sqrt{2\eta}\,\boldsymbol{\epsilon}_{j},
\end{aligned}
\end{equation}
where $\eta$ is the step size and $\boldsymbol{\epsilon}_j \sim \mathcal{N}(0, \mathbf{I})$.
In standard MCMC, the prior gradient $\nabla_{\mathbf{x}_0}\log p(\mathbf{x}_0)$ serves as a regularization term.
However, because the diffusion initialization places $\hat{\mathbf{x}}_0$ in a region where $\|\nabla\log p(\mathbf{x}_0)\| = O(\varepsilon)$, this contribution is numerically negligible.
Once we abandon the time-marginal requirement entirely, convergence is governed purely by the geometry of the likelihood, and far fewer steps suffice.
We therefore omit the prior gradient and obtain a likelihood-only refinement:
\begin{equation}
\mathbf{x}_0^{(j+1)}
=
\mathbf{x}_0^{(j)}
- \frac{\eta}{\gamma^2}\,
  \nabla_{\mathbf{x}_0^{(j)}}
  \left\|\,\mathbf{y} - \mathcal{A}\!\left(\mathbf{x}_0^{(j)}\right)\right\|^{2}
+ \sqrt{2\eta}\,\boldsymbol{\epsilon}_{j}.
\label{eq:mcmc_gaussian}
\end{equation}
This refinement drives the estimate toward the measurement-consistent solution but
reduces variability in noise-corrupted or unobserved components.
To restore this variability, each refinement cycle is followed by a re-noising step.
This controlled stochasticity prevents overfitting and allows the next cycle to explore
from a fresh state while maintaining measurement consistency.
Unlike maximum a posteriori (MAP) estimation, our method performs stochastic sampling
in the diffusion-defined latent space, exploring the posterior rather than collapsing
to one mode.
To improve convergence stability, we adopt an EDM-inspired annealing schedule in which the step size decays polynomially (exponent in $[-4, -7]$), concentrating more iterations in the low-noise regime where fine-scale refinement is most effective.

To provide a comprehensive structural overview of the proposed framework, we present the complete pseudocode for DAPS++ in Algorithm 1. This outlines the heterogeneous decomposition strategy, transitioning from the prior-dominant generation in Stage~1 to the likelihood-dominant refinement in Stage~2.


\begin{figure}[t]
\centering
\begin{tabular}{@{}p{0.9\textwidth}@{}}
\toprule
\textbf{Algorithm 1} DAPS++ Inference with Decoupled Initialization-Refinement \\
\midrule
\textbf{Require:} Measurement $\mathbf{y}$, operator $\mathcal{A}$, effective noise variance $\gamma^2$; diffusion score $s_\theta$, noise schedule $\{\sigma_t\}$, threshold $\bar{\sigma}$, refinement steps $J$, step sizes $\{\eta_j\}$, total cycles $K$. \\
\textbf{Ensure:} Final reconstruction $\hat{\mathbf{x}}_0$. \\
1: Initialize $x_{\mathrm{in}}^{(0)} \sim \mathcal{N}(\mathbf{0},\sigma_{\max}^2\mathbf{I})$ \\
2: \textbf{for} $k = 1$ \textbf{to} $K$ \textbf{do} \hfill \textcolor{gray}{$\triangleright$ Stage 1: Diffusion initialization} \\
3: \hspace{1em} \textbf{if} $\sigma_{\mathrm{in}} > \bar{\sigma}$ \textbf{then} \\
4: \hspace{2.5em} $\hat{\mathbf{x}}_0^{(k)} \leftarrow x_{\mathrm{in}}^{(k-1)} + \sigma_{\mathrm{in}}^{\,2} s_\theta\!\big(x_{\mathrm{in}}^{(k-1)},\sigma_{\mathrm{in}}\big)$ \\
5: \hspace{1em} \textbf{else} \\
6: \hspace{2.5em} $\hat{\mathbf{x}}_0^{(k)} \leftarrow \mathtt{ODESolver}\big(x_{\mathrm{in}}^{(k-1)},\sigma_{\mathrm{in}}\big)$ \\
7: \hspace{1em} \textbf{end if} \hfill \textcolor{gray}{$\triangleright$ Stage 2: Likelihood-driven refinement} \\
8: \hspace{1em} $z^{(0)} \leftarrow \hat{\mathbf{x}}_0^{(k)}$ \\
9: \hspace{1em} \textbf{for} $j = 1$ \textbf{to} $J$ \textbf{do} \\
10: \hspace{2.5em} $\mathbf{r}^{(j-1)} \leftarrow \mathbf{y} - \mathcal{A}(z^{(j-1)})$ \\
11: \hspace{2.5em} $\nabla_z \mathcal{L} \leftarrow \frac{1}{\gamma^2} \mathtt{VJP}_{\mathcal{A}}\big(z^{(j-1)},\mathbf{r}^{(j-1)}\big)$ \\
12: \hspace{2.5em} Sample $\boldsymbol{\xi}_j \sim \mathcal{N}(\mathbf{0},\mathbf{I})$ \\
13: \hspace{2.5em} $z^{(j)} \leftarrow z^{(j-1)} + \eta_j \nabla_z \mathcal{L} + \sqrt{2\eta_j}\,\boldsymbol{\xi}_j$ \hfill \textcolor{gray}{$\triangleright$ ULA refinement} \\
14: \hspace{1em} \textbf{end for} \\
15: \hspace{1em} $\tilde{\mathbf{x}}_0^{(k)} \leftarrow z^{(J)}$ \hfill \textcolor{gray}{$\triangleright$ Re-noising for next iteration} \\
16: \hspace{1em} Sample $\boldsymbol{\epsilon}_k \sim \mathcal{N}(\mathbf{0},\mathbf{I})$ \\
17: \hspace{1em} $x_{\mathrm{in}}^{(k)} \leftarrow \tilde{\mathbf{x}}_0^{(k)} + \sigma_{k}\,\boldsymbol{\epsilon}_k$ \\
18: \textbf{end for} \\
19: \textbf{return} $\hat{\mathbf{x}}_0 \leftarrow \tilde{\mathbf{x}}_0^{(K)}$ \\
\bottomrule
\end{tabular}
\end{figure}

\subsection{Initialization Quality and Convergence Analysis}
\label{sec:method:3}
To justify this decoupled strategy, we analyze the initialization quality and refinement convergence in turn, showing that once the estimate is near the data manifold, each stage needs to be performed only once.

\textbf{Initialization quality.} At a given noise level $\sigma_{t}$, the Tweedie estimate has already recovered the dominant structure of the signal from the score function: large-scale geometry, color distribution, and semantic content are established. The measurement residual at this point is of order $O(|\mathcal{A}|\sigma_{t}) + O(\gamma\sqrt{d_y})$, where $d_y$ is the measurement dimension. At early annealing stages where $\sigma_{t}$ is large, the residual is also large, but high precision is not required since re-noising at the end of each cycle erases fine-scale corrections; at late stages where $\sigma_{t}$ is small, the residual has shrunk to near the noise floor and only a few Langevin steps are needed. With $\bar{\sigma} = 10\gamma$, the transition from Tweedie to a single RK4 step occurs when the first-order truncation error begins to dominate the residual.

\textbf{Convergence across problem types.} The number of refinement steps required in Stage~2 is governed by two properties of the measurement operator in the neighborhood of $\hat{\mathbf{x}}_0$. The first is the condition number of the forward map, which controls the convergence rate of the likelihood gradient. The second is the degree of convexity of the negative log-likelihood near the initialization, which determines whether local refinement can reach the correct solution. A formal convergence bound relating these quantities to the required step count is provided in the Appendix. For well-conditioned linear operators, the likelihood is quadratic and convergence is fast: super-resolution requires only 2 steps because the downsampling operator preserves most signal energy, while Gaussian deblurring requires more steps due to its broader eigenvalue spread. For mild nonlinearities such as per-pixel tone mapping in HDR, the likelihood remains approximately convex near the initialization and convergence is comparable to the linear case (5 steps). For stronger nonlinearities such as nonlinear deblurring, the signal-dependent kernel introduces inhomogeneous curvature, requiring more iterations (50 steps). In phase retrieval, the likelihood has multiple equivalent minima; if the initialization falls in the wrong basin, local refinement cannot recover the correct solution.

\subsection{Connections to Existing Diffusion-Based Solvers}
\label{sec:method:4}
The two-stage formulation clarifies why DAPS++ achieves its efficiency gains through structural decoupling rather than parameter reduction.
Although many diffusion-based inverse problem approaches are presented as posterior samplers, recent studies show that their updates often resemble MAP-style iterative refinements~\cite{xu2025rethinking}.
The distinction between methods lies in how tightly they couple the diffusion prior with the likelihood at each step, and what computational cost that coupling incurs.

In DPS~\cite{chung2022diffusion}, the data-consistency term requires differentiating
the measurement residual $\|\mathbf{y} - \mathcal{A}(\hat{\mathbf{x}}_0(\mathbf{x}_t))\|_2^2$
through Tweedie's estimator $\hat{\mathbf{x}}_0$, propagating gradients back through
the score function at every diffusion timestep.
This Jacobian-vector product through the score function at each of $T$ steps is the
primary computational bottleneck of DPS, and makes it structurally incompatible with
higher-order ODE solvers, where additional score evaluations per step would each
require a corresponding backward pass.
From~\cref{eq:inner_prod}, the prior and likelihood gradients are statistically
independent, so the DPS update decomposes as
\begin{equation}
\begin{aligned}
\mathbf{x}_{t-1}'
&= \underbrace{
     \mathbf{x}_t
     + \sigma_t^{2}\, s_\theta(\mathbf{x}_t,\sigma_t)
     + \sigma_{t-1}\mathbf{z}
   }_{\text{initialization}} \\
&\quad
  - \underbrace{
      \eta_t
      \Big[
          \nabla_{\mathbf{x}_0}\,\|\mathbf{y} - \mathcal{A}(\mathbf{x}_0)\|_2^{2}
      \Big]_{\mathbf{x}_0 = \hat{\mathbf{x}}_0(\mathbf{x}_t)}
    }_{\text{refinement}},
\end{aligned}
\label{eq:dps_decomp}
\end{equation}
where $\mathbf{z} \sim \mathcal{N}(\mathbf{0},\mathbf{I})$ and $\eta_t$ denotes the step size.
DPS therefore performs initialization and a single likelihood correction at every timestep, but couples them so that the score network and measurement operator must be evaluated jointly at each step.

DAPS~\cite{zhang2025improving} breaks this coupling at the algorithmic level by alternating Tweedie estimation with Langevin dynamics. By applying this directly to $\hat{\mathbf{x}}_0$ in image space, the method iteratively updates the clean-image estimate without backpropagating through the score function. However, because DAPS re-invokes the score network at each annealing level to preserve the time-marginal distribution $p(\mathbf{x}_t|\mathbf{y})$, the diffusion prior remains an active ingredient throughout sampling. As a result, a large number of Langevin steps per annealing level are required to approximately sample from $p(\mathbf{x}_0|\mathbf{x}_t, \mathbf{y})$, which is the necessary condition for theoretically preserving the time-marginal $p(\mathbf{x}_t|\mathbf{y})$ at each level. Without sufficiently converged Langevin samples at each stage, this guarantee breaks down.

DAPS++ takes this separation to its logical conclusion: the score network is used only in Stage~1, and Stage~2 operates entirely under the likelihood with no further access to the prior, removing the time-marginal constraint and yielding the approximately 90\% reduction in NFEs reported in Sec.~\ref{exp:sec:abla}.

\section{Experiments}
\label{sec:exp}

\begin{figure}[ht]
  \centering
    \includegraphics[width=\linewidth, height=0.5\textheight, keepaspectratio]{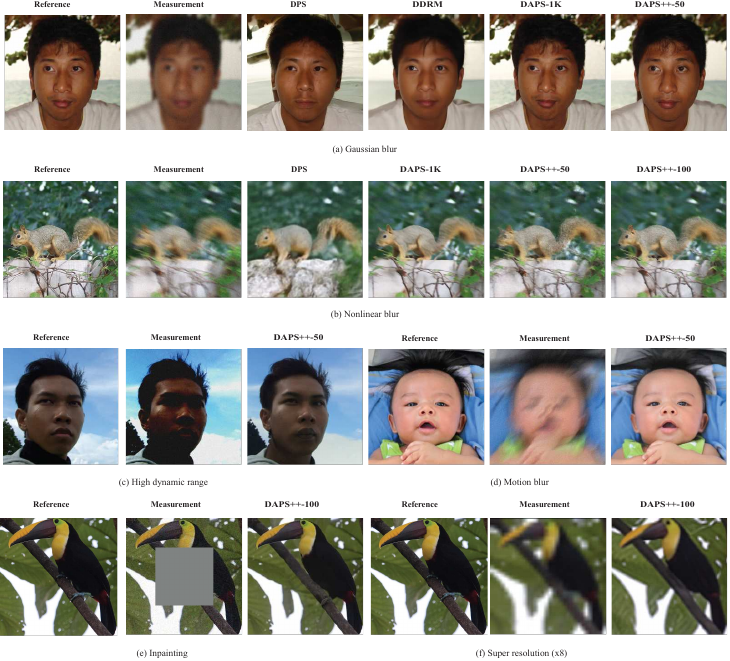}
\caption{(a)~Gaussian blur reconstruction on the FFHQ-256 dataset compared with selected baselines.
(b)~Nonlinear blur reconstruction on ImageNet, comparing \textbf{DAPS++-50} and \textbf{DAPS++-100} alongside selected baselines.
(c)~High dynamic range (HDR) reconstruction on the FFHQ-256 dataset.
(d)~Motion blur reconstruction on the FFHQ-256 dataset.
(e)~Box inpainting results on ImageNet, where \textbf{DAPS++} accurately restores structural details.
(f)~$8\times$ super-resolution using a pre-trained ImageNet diffusion model, demonstrating strong visual fidelity.}
  \label{fig:qaulitive}
\end{figure}

\subsection{Experimental Setup}
We evaluate our method using pixel-space pre-trained diffusion models on the FFHQ~\cite{chung2022diffusion} and ImageNet~\cite{dhariwal2021diffusion} datasets.
For the noise schedule, we adopt the POLY(-7) discretization from EDM~\cite{karras2022elucidating}.
To improve efficiency, $\hat{\mathbf{x}}_0(\mathbf{x}_t)$ is generated using~\cref{eq:Tweedie} while $\sigma_t > \bar{\sigma}$, and a single RK4 update is used once $\sigma_t \le \bar{\sigma}$.
Empirically, we set $\bar{\sigma} = 10\times\gamma$ to balance computational cost and reconstruction accuracy.
The total number of neural function evaluations (NFEs) remains low because each iteration requires only one network evaluation, further reduced by an annealed step-size schedule.

\begin{figure}[t]
  \centering
  \includegraphics[width=0.85\linewidth]{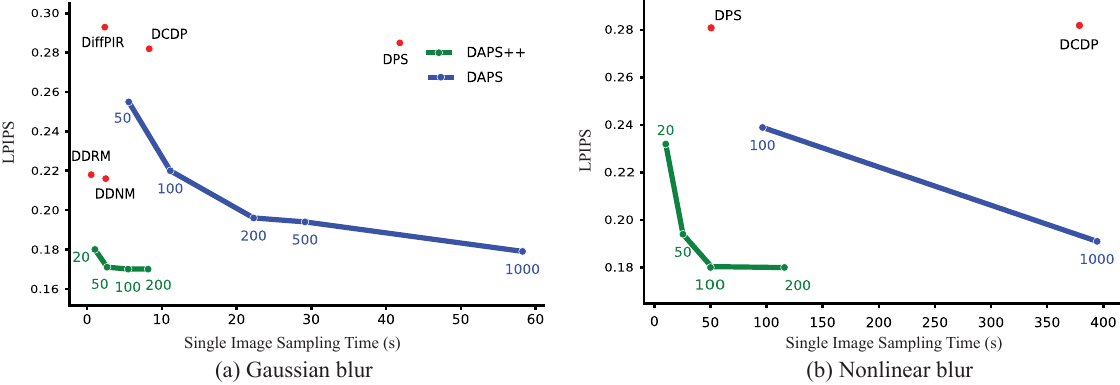}
  \caption{Sampling time vs.\ reconstruction quality on an NVIDIA A100 (80GB PCIe) GPU.
  The y-axis shows LPIPS; results are averaged over 100 FFHQ images for Gaussian blur and nonlinear blur.}
  \label{fig:time_compare}
\end{figure}

We use DAPS++-50 (50 annealing steps) for FFHQ and DAPS++-100 (100 steps) for ImageNet.
Each DAPS++ iteration performs 4--8 MCMC refinement steps, except for nonlinear deblurring, which requires around 50 refinement steps in the refinement stage.
Learning rates are tuned individually for each task.
Additional details regarding model configurations, samplers, and hyperparameters appear in the Appendix, along with a discussion of sampling efficiency in~\cref{fig:time_compare}.

Because the diffusion model already captures the underlying structure and fine-scale statistics of the data distribution, it effectively denoises the input while restoring plausible high-frequency details.
Since the diffusion output approximates $\mathcal{A}(\mathbf{x}_0)$, the residual between this prediction and the measurement $\mathbf{y}$ is mostly dominated by noise.
To ensure stable convergence and avoid overfitting to measurement noise, the injected diffusion noise at the end of each iteration must be smaller than the additive noise level.
In practice, we use a noise schedule decaying from $\sigma_{\max}=100$ to $\sigma_{\min}=0.1$ with measurement noise $\gamma=0.05$, which effectively suppresses overfitting and preserves structural fidelity throughout the DAPS family.

We compare our method against several state-of-the-art diffusion-based inverse problem solvers, including DAPS-1K, DAPS-100~\cite{zhang2025improving}, DPS~\cite{chung2022diffusion}, DDRM~\cite{kawar2022DDRM}, DDNM~\cite{kawar2022DDRM}, DiffPIR~\cite{zhu2023denoising}, and DCDP~\cite{li2024decoupled}.
Note that SVD-based methods such as DDRM and DDNM apply only to linear inverse problems.

\textbf{Forward measurement operators.}
We evaluate our approach on a variety of linear and nonlinear inverse problems. Linear tasks include super-resolution, Gaussian and motion deblurring, and inpainting with box or random masks. For Gaussian and motion blur, we use $61\times61$ kernels with standard deviations of 3.0 and 0.5, respectively. Super-resolution employs bicubic downsampling by a factor of 4, while inpainting tasks use the standard $128\times128$ box mask. Nonlinear tasks include high dynamic range (HDR) reconstruction and nonlinear deblurring, following configurations described in~\cite{zhang2025improving}.\footnote{Frequency-domain-based phase retrieval is discussed in the Supplementary Material, Section~S2.2.} All measurements are corrupted by additive white noise with standard deviation $\gamma = 0.05$.

\textbf{Datasets and metrics.}
We evaluate the proposed method on two benchmark datasets: FFHQ ($256\times256$)~\cite{karras2019style} and ImageNet ($256\times256$)~\cite{deng2009imagenet}.
For both datasets, 100 validation images are used for quantitative evaluation under identical conditions.
To comprehensively assess reconstruction quality, we report SSIM~\cite{wang2004image}, LPIPS~\cite{zhang2018unreasonable}, and FID~\cite{heusel2017gans} as our primary metrics.
All methods---including ours and the baselines---are evaluated on images normalized to $[0,1]$ to ensure consistency and comparability across datasets.

\subsection{Results}

\begin{table}[tb]
\caption{Quantitative comparison on 100 validation images of \textbf{FFHQ} across four linear and two nonlinear inverse problems.
Metrics are reported as SSIM$\uparrow$ / LPIPS$\downarrow$ / FID$\downarrow$.
Bold indicates the best result; underlined values denote the second best.
All measurements include additive Gaussian noise with $\gamma = 0.05$.}
\label{tab:ffhq}
\centering
\renewcommand{\arraystretch}{1.3}
\resizebox{\linewidth}{!}{
\begin{tabular}{@{}lcccccc@{}}
\toprule
Method &
SR $\times$4 &
Inpainting &
Gaussian Blur &
Motion Blur &
Nonlinear Blur &
HDR \\
\midrule
 & SSIM / LPIPS / FID &
   SSIM / LPIPS / FID &
   SSIM / LPIPS / FID &
   SSIM / LPIPS / FID &
   SSIM / LPIPS / FID &
   SSIM / LPIPS / FID \\
\midrule
DDNM & 0.720 / 0.290 / 152.9 & \underline{0.810} / \underline{0.162} / 94.7 & \textbf{0.804} / 0.216 / 57.8 & --- & --- & --- \\
DDRM & \textbf{0.820} / \underline{0.191} / 76.1 & 0.792 / 0.210 / 88.9 & \underline{0.786} / 0.218 / 89.4 & --- & --- & --- \\
DiffPIR & 0.545 / 0.269 / 60.0 & 0.525 / 0.344 / 63.8 & 0.514 / 0.293 / 65.3 & 0.553 / 0.257 / 52.6 & --- & --- \\
DCDP & 0.642 / 0.333 / 92.9 & 0.757 / 0.167 / 42.3 & 0.719 / 0.282 / 82.2 & 0.508 / 0.384 / 115.2 & \textbf{0.803} / \textbf{0.160} / \textbf{44.8} & --- \\
DPS & 0.591 / 0.357 / 81.1 & 0.727 / 0.259 / 75.6 & 0.647 / 0.285 / 73.3 & 0.588 / 0.327 / 76.8 & 0.648 / 0.281 / 74.3 & 0.693 / 0.284 / 77.6 \\
\midrule
DAPS-100 & 0.721 / 0.233 / 69.3 & 0.716 / 0.194 / 52.9 & 0.731 / 0.220 / 63.4 &
0.766 / 0.182 / 54.8 & 0.685 / 0.239 / 68.1 & 0.819 / 0.184 / 44.9 \\
DAPS-1K & \underline{0.782} / 0.192 / \underline{55.5} &
0.747 / 0.176 / \underline{50.1} &
\underline{0.786} / \underline{0.179} / \underline{52.7} &
\textbf{0.836} / \underline{0.137} / \underline{38.4} &
\underline{0.762} / \underline{0.191} / \underline{57.8} &
\textbf{0.839} / \textbf{0.163} / \textbf{41.1} \\
\textbf{DAPS++-50} & 0.781 / \textbf{0.176} / \textbf{46.0} &
\textbf{0.812} / \textbf{0.141} / \textbf{42.1} &
0.784 / \textbf{0.171} / \textbf{51.1} &
\underline{0.829} / \textbf{0.136} / \textbf{37.9} &
0.745 / 0.194 / 54.9 &
\underline{0.834} / \underline{0.169} / \underline{42.3} \\
\bottomrule
\end{tabular}}
\end{table}
\begin{table}[tb]
\caption{Quantitative comparison on 100 validation images of \textbf{ImageNet} with four linear and two nonlinear tasks.
Metrics follow SSIM$\uparrow$ / LPIPS$\downarrow$ / FID$\downarrow$.
Bold values denote the best performance; underlined entries indicate the second best.
Each measurement includes additive Gaussian noise with $\gamma = 0.05$.}
\label{tab:imagenet}
\centering
\renewcommand{\arraystretch}{1.3}
\resizebox{\linewidth}{!}{
\begin{tabular}{@{}lcccccc@{}}
\toprule
Method &
SR $\times$4 &
Inpainting &
Gaussian Blur &
Motion Blur &
Nonlinear Blur &
HDR \\
\midrule
 & SSIM / LPIPS / FID &
  SSIM / LPIPS / FID &
  SSIM / LPIPS / FID &
  SSIM / LPIPS / FID &
  SSIM / LPIPS / FID &
  SSIM / LPIPS / FID \\
\midrule
DDNM & 0.487 / 0.458 / 215.3 & 0.656 / 0.289 / 157.9 & 0.650 / \textbf{0.231} / \textbf{90.8} & --- & --- & --- \\
DDRM & 0.652 / 0.296 / 111.2 & 0.602 / 0.336 / 163.7 & 0.584 / 0.361 / 161.0 & --- & --- & --- \\
DiffPIR & 0.443 / 0.370 / \textbf{106.8} & 0.500 / 0.390 / 149.2 & 0.395 / 0.428 / 154.6 & 0.152 / 0.683 / 264.8 & --- & --- \\
DPS & 0.454 / 0.474 / 208.4 & 0.623 / 0.343 / 157.5 & 0.526 / 0.355 / 155.6 & 0.542 / 0.365 / 161.4 & 0.520 / 0.365 / 151.5 & 0.349 / 0.552 / 213.8 \\
\midrule
DAPS-100 & 0.610 / 0.309 / 114.4 & 0.693 / 0.248 / 123.6 & 0.619 / 0.296 / 117.0 &
0.710 / 0.209 / 62.0 & 0.650 / 0.263 / 105.7 & \underline{0.810} / \underline{0.189} / \textbf{43.2} \\
DAPS-1K & 0.638 / \underline{0.295} / \underline{109.1} &
0.715 / 0.229 / 114.0 &
0.658 / \underline{0.268} / \underline{107.6} &
\textbf{0.769} / \textbf{0.175} / \textbf{47.1} &
\textbf{0.720} / \underline{0.212} / \underline{75.9} &
\textbf{0.824} / \textbf{0.171} / \underline{44.7} \\
\textbf{DAPS++-50} & \underline{0.653} / \underline{0.283} / 111.7 &
\underline{0.760} / \underline{0.208} / \underline{113.3} &
\textbf{0.666} / 0.269 / 118.0 &
0.754 / 0.185 / 57.4 &
0.686 / 0.234 / 85.5 &
0.807 / \underline{0.189} / 46.9 \\
\textbf{DAPS++-100} & \textbf{0.661} / \textbf{0.276} / 114.2 &
\textbf{0.771} / \textbf{0.195} / \textbf{105.3} &
\underline{0.663} / 0.273 / 122.9 &
\underline{0.763} / \underline{0.179} / \underline{56.1} &
\underline{0.718} / \textbf{0.211} / \textbf{75.7} &
0.807 / 0.191 / 48.5 \\
\bottomrule
\end{tabular}}
\end{table}

\textbf{Main results.}
\cref{tab:ffhq,tab:imagenet} present the quantitative comparisons against various baselines, with qualitative examples shown in~\cref{fig:qaulitive}.
Across both linear and nonlinear inverse problems, our method achieves competitive---or superior---reconstruction quality with higher SSIM and lower LPIPS and FID, while requiring substantially fewer NFEs than DAPS-1K.
Under identical NFE budgets, DAPS++ consistently attains higher fidelity, demonstrating the effectiveness of fully decoupling the prior and likelihood stages and avoiding overfitting in either stage.
DAPS++ also shows strong robustness under heavily degraded conditions, such as bicubic $8\times$ super-resolution, where it can still recover meaningful structural information using only the ImageNet pre-trained model.

During the alternating initialization-refinement cycles, the optimization stages are deliberately non-overlapping.
Conventional MAP-style updates in the refinement stage typically overfit measurement noise unless carefully regularized.
In contrast, DAPS++ employs larger step sizes and fewer refinement iterations, leveraging the structure provided by the diffusion-based initialization while avoiding noise amplification.
This design yields stable convergence and improved reconstruction quality at lower computational cost.

Small-step ODE solvers introduce notable discretization error when only a limited number of updates are performed.
For example, DAPS-1K uses five Euler steps, resulting in an $\mathcal{O}(\Delta t)$ error that produces a weaker initialization compared with Tweedie-based estimation.
As a consequence, more MCMC iterations are needed to return to the posterior manifold.
DAPS++, by directly predicting the denoised mean in the initialization stage, avoids this discretization bottleneck and converges more rapidly with significantly fewer total steps.
All baseline methods are evaluated on images normalized to $[0,1]$ to ensure consistency and comparability across datasets.

\begin{figure}[t]
  \centering
  \includegraphics[width=0.85\linewidth]{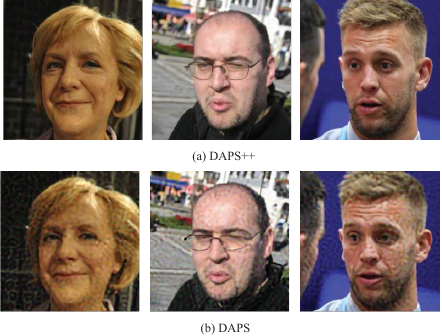}
  \caption{Example under $\gamma{=}0.1$, $\sigma_{\min}{=}0.1$.
  Comparison between (a)~\textbf{DAPS++} and (b)~\textbf{DAPS} for Gaussian blur shows that DAPS overfits noise, producing more artifacts in the final outputs.}
  \label{fig:noise}
\end{figure}

\textbf{Noise Tolerance.}
DAPS++ exhibits strong robustness under increasing noise levels.
As shown in~\cref{tab:noise,fig:noise}, when $\sigma_{\min}=0.1$ and $\gamma=0.1$, DAPS tends to overfit the noise, producing visible artifacts.
In contrast, DAPS++ preserves structural fidelity and perceptual quality even as the noise level increases.
As $\gamma$ rises from 0.05 to 0.4, DAPS++ consistently exhibits stronger robustness, especially when $\sigma_{\min}$ is matched to $\gamma$, demonstrating its ability to prevent overfitting.
This robustness arises from the structured re-noising mechanism and the reduced number of refinement steps, which together prevent overfitting and maintain diffusion-driven variability.

\subsection{Ablation Studies}
\label{exp:sec:abla}
\textbf{Efficiency comparison.}
We assess the efficiency of DAPS++ under varying computational budgets by evaluating configurations with diffusion NFEs ranging from 20 to 200, and comparing them against DAPS variants using 50 to 1K NFEs as well as several representative baselines. All methods are evaluated in terms of single-image sampling time on an NVIDIA A100 (80GB PCIe) GPU and reconstruction quality using the LPIPS metric. As shown in~\cref{fig:time_compare}, the trade-off between sampling cost and reconstruction quality consistently favors DAPS++, which achieves higher fidelity at substantially lower computational cost across both linear and nonlinear inverse problems. Even at extremely low computational settings, such as DAPS++--20, the method converges rapidly and produces competitive reconstructions for linear degradation models, suggesting that the fully decoupled formulation provides an effective path toward stable likelihood-driven refinement without relying on heavy diffusion sampling. As computational budgets increase, DAPS++ exhibits progressively improved performance while maintaining a significantly lower cost profile than diffusion-based alternatives that interleave prior and likelihood updates at every timestep. Overall, these results demonstrate that DAPS++ achieves a favorable balance between accuracy and efficiency, making it a scalable and practical solution for real-time or resource-limited inverse imaging applications.

\begin{table}[t]
  \caption{Comparison of \textbf{DAPS-1K} and \textbf{DAPS++-50} under varying noise levels on the FFHQ Gaussian blur task, where $\sigma_{\min}$ is matched to $\gamma$.}
  \label{tab:noise}
  \centering
  \renewcommand{\arraystretch}{1.4}
  \resizebox{\linewidth}{!}{
  \begin{tabular}{lcccc}
  \toprule
  \textbf{Noise Level} &
  $\gamma{=}0.05$, $\sigma_{\min}{=}0.1$ &
  $\gamma{=}0.1$, $\sigma_{\min}{=}0.1$ &
  $\gamma{=}0.2$, $\sigma_{\min}{=}0.2$ &
  $\gamma{=}0.4$, $\sigma_{\min}{=}0.4$ \\
  \cmidrule(lr){2-5}
  & SSIM$\uparrow$ / LPIPS$\downarrow$ &
    SSIM$\uparrow$ / LPIPS$\downarrow$ &
    SSIM$\uparrow$ / LPIPS$\downarrow$ &
    SSIM$\uparrow$ / LPIPS$\downarrow$ \\
  \midrule
  DAPS-1K &
  0.786 / 0.179 &
  0.615 / 0.330 &
  0.410 / 0.495 &
  0.192 / 0.633 \\
  \textbf{DAPS++-50} &
  \textbf{0.784} / \textbf{0.171} &
  \textbf{0.728} / \textbf{0.191} &
  \textbf{0.674} / \textbf{0.229} &
  \textbf{0.579} / \textbf{0.303} \\
  \bottomrule
  \end{tabular}}
\end{table}

\textbf{Comparison with DAPS on prior and data-consistency design.}
We examine the key design difference between DAPS++ and DAPS by evaluating how the diffusion prior is used during reconstruction. DAPS applies diffusion score together with likelihood updates and requires a large number of MCMC steps to approximately preserve the assumed time-marginal distribution. In contrast, DAPS++ uses prior guidance only in the initialization stage, after which a small number of likelihood-only refinement steps is sufficient to enforce data consistency.

Tab.~\ref{tab:dapspp_ablation_inpainting} shows the results of a focused FFHQ inpainting ablation under the same 100 diffusion-NFE budget. Comparing DAPS-100 with and without prior-gradient refinement shows nearly unchanged image quality and perceptual metrics, suggesting that the score gradient from diffusion guidance contributes little during data refinement in this setting. In contrast, the initialization strategy has a clearer effect. Direct Tweedie initialization improves the metrics over the DAPS baseline but tends to produce overly smooth reconstructions with insufficient details, motivating our thresholded transition from Tweedie estimation to a higher-order solver for initialization. Under the same computation budget, DAPS++ achieves better reconstruction quality and lower runtime than the DAPS baseline, indicating that the roles and design of the diffusion prior and data-consistency term are critical.

\begin{table}[t]
    \caption{
    FFHQ inpainting ablation under the same 100 diffusion-NFE budget, isolating the effects of initialization and refinement strategy.
    }
  \label{tab:dapspp_ablation_inpainting}
  \centering
  \renewcommand{\arraystretch}{1.4}
  \resizebox{\linewidth}{!}{%
  \begin{tabular}{lcccc}
  \toprule
  \textbf{Method} &
  \textbf{Initialization} &
  \textbf{Refinement} &
  SSIM$\uparrow$ / LPIPS$\downarrow$ &
  \textbf{Time (s)} \\
  \midrule
  DAPS-100 &
  2-step Euler &
  Prior + likelihood &
  0.716 / 0.194 &
  7.4 \\
  DAPS-100 w/o prior-gradient &
  2-step Euler &
  Likelihood only &
  0.716 / 0.194 &
  6.8 \\
  DAPS + Tweedie init. &
  Tweedie &
  Prior + likelihood &
  0.741 / 0.186 &
  10.1 \\
  DAPS + Tweedie/RK4 &
  Tweedie/RK4 &
  Prior + likelihood &
  0.712 / 0.192 &
  5.9 \\
  \textbf{DAPS++} &
  Tweedie/RK4 &
  Likelihood only &
  \textbf{0.812} / \textbf{0.141} &
  \textbf{2.4} \\
  \bottomrule
  \end{tabular}%
  }
\end{table}

\textbf{Hyperparameter choice.}
Two hyperparameters play a central role in our method: the threshold noise level $\bar{\sigma}$ and the polynomial scheduler exponent. Both parameters directly influence reconstruction quality, particularly when the number of sampling steps is limited and the interaction between diffusion initialization and likelihood refinement becomes more sensitive. The threshold $\bar{\sigma}$ determines when Tweedie-based estimation transitions to a higher-order ODE solver within the diffusion process, allowing efficient denoising at large noise levels while producing a reliable initialization with only a few high-order ODE evaluations. However, even under small noise settings, purely diffusion-generated estimates without data-driven refinement will drift away from the measurement; therefore, after the higher-order solver step, we still apply a refinement update at the end of the annealing schedule to correct this deviation.

The polynomial scheduler exponent governs the decay rate of the noise levels during annealing and becomes especially important in low-step regimes (\eg, 20--50 steps), as it controls the trade-off between convergence stability and refinement strength. A slower decay stabilizes the refinement updates but introduces more high-noise transitions, which may reduce the effectiveness of likelihood refinement. A faster decay generally accelerates convergence and is often preferred in practice, provided that the refinement stage remains numerically stable. A comprehensive sensitivity analysis for these hyperparameters, along with practical tuning guidelines, is provided in the Appendix.

\section{Conclusion}
\label{sec:con}
In this work, we revisit the underlying mechanism of diffusion-based inverse problem solvers and propose a fully decoupled initialization-refinement formulation that makes explicit the separation between the diffusion and data-consistency stages.
Based on this perspective, we develop \textbf{DAPS++}, a fully decoupled framework that substantially improves sampling efficiency while preserving high reconstruction fidelity.
The diffusion stage functions as a prior-driven initialization on a flat manifold, whereas the data-consistency stage performs likelihood-guided refinement independent of the diffusion process.
This separation offers a clearer functional interpretation of diffusion-based inference and provides an explanation for the empirical behavior of existing approaches.
Across both linear and nonlinear imaging tasks, DAPS++ achieves strong performance using only a fraction of the function evaluations required by conventional methods, establishing an efficient and scalable foundation for future research in diffusion-based inverse problems.

%
%
\clearpage
\bibliographystyle{splncs04}
\bibliography{main}

@String(ICCV= {Int. Conf. Comput. Vis.})

@String(ICASSP=	{ICASSP})

@String(ICLR = {Int. Conf. Learn. Represent.})

@String(ICCV  = {ICCV})

@String(ICLR  = {ICLR})

@article{lucy1994optimum,
  title={Optimum strategies for inverse problems in statistical astronomy},
  author={Lucy, LB},
  journal={Astronomy and Astrophysics (ISSN 0004-6361), vol. 289, no. 3, p. 983-994},
  volume={289},
  pages={983--994},
  year={1994}
}

@inproceedings{song2021solving,
  title={Solving inverse problems in medical imaging with score-based generative models},
  author={Song, Yang and Shen, Liyue and Xing, Lei and Ermon, Stefano},
  booktitle={International Conference on Learning Representations},
  year={2022}
}

@article{dey2006richardson,
  title={Richardson--Lucy algorithm with total variation regularization for 3D confocal microscope deconvolution},
  author={Dey, Nicolas and Blanc-Feraud, Laure and Zimmer, Christophe and Roux, Pascal and Kam, Zvi and Olivo-Marin, Jean-Christophe and Zerubia, Josiane},
  journal={Microscopy research and technique},
  volume={69},
  number={4},
  pages={260--266},
  year={2006},
  publisher={Wiley Online Library}
}

@article{porth2019event,
  title={The event horizon general relativistic magnetohydrodynamic code comparison project},
  author={Porth, Oliver and Chatterjee, Koushik and Narayan, Ramesh and Gammie, Charles F and Mizuno, Yosuke and Anninos, Peter and Baker, John G and Bugli, Matteo and Chan, Chi-kwan and Davelaar, Jordy and others},
  journal={The Astrophysical Journal Supplement Series},
  volume={243},
  number={2},
  pages={26},
  year={2019},
  publisher={IOP Publishing}
}

@article{fish1995blind,
  title={Blind deconvolution by means of the Richardson--Lucy algorithm},
  author={Fish, DA and Brinicombe, AM and Pike, ER and Walker, JG},
  journal={Journal of the Optical Society of America A},
  volume={12},
  number={1},
  pages={58--65},
  year={1995},
  publisher={Optical Society of America}
}

@article{lustig2007sparse,
  title={Sparse MRI: The application of compressed sensing for rapid MR imaging},
  author={Lustig, Michael and Donoho, David and Pauly, John M},
  journal={Magnetic Resonance in Medicine: An Official Journal of the International Society for Magnetic Resonance in Medicine},
  volume={58},
  number={6},
  pages={1182--1195},
  year={2007},
  publisher={Wiley Online Library}
}

@article{stuart2010inverse,
  title={Inverse problems: a Bayesian perspective},
  author={Stuart, Andrew M},
  journal={Acta numerica},
  volume={19},
  pages={451--559},
  year={2010},
  publisher={Cambridge University Press}
}

@incollection{dashti2015bayesian,
  title={The Bayesian approach to inverse problems},
  author={Dashti, Masoumeh and Stuart, Andrew M},
  booktitle={Handbook of uncertainty quantification},
  pages={1--118},
  year={2015},
  publisher={Springer}
}

@article{ho2020denoising,
  title={Denoising diffusion probabilistic models},
  author={Ho, Jonathan and Jain, Ajay and Abbeel, Pieter},
  journal={Advances in neural information processing systems},
  volume={33},
  pages={6840--6851},
  year={2020}
}

@article{song2019generative,
  title={Generative modeling by estimating gradients of the data distribution},
  author={Song, Yang and Ermon, Stefano},
  journal={Advances in neural information processing systems},
  volume={32},
  year={2019}
}

@article{song2020improved,
  title={Improved techniques for training score-based generative models},
  author={Song, Yang and Ermon, Stefano},
  journal={Advances in neural information processing systems},
  volume={33},
  pages={12438--12448},
  year={2020}
}

@inproceedings{song2021scorebased,
  title={Score-based generative modeling through stochastic differential equations},
  author={Song, Yang and Sohl-Dickstein, Jascha and Kingma, Diederik P and Kumar, Abhishek and Ermon, Stefano and Poole, Ben},
  booktitle=ICLR,
  year={2021}
}

@article{karras2022elucidating,
  title={Elucidating the design space of diffusion-based generative models},
  author={Karras, Tero and Aittala, Miika and Aila, Timo and Laine, Samuli},
  journal={Advances in neural information processing systems},
  volume={35},
  pages={26565--26577},
  year={2022}
}

@article{gibbs2002choosing,
  title={On choosing and bounding probability metrics},
  author={Gibbs, Alison L and Su, Francis Edward},
  journal={International statistical review},
  volume={70},
  number={3},
  pages={419--435},
  year={2002},
  publisher={Wiley Online Library}
}

@inproceedings{song2020denoising,
  title={Denoising diffusion implicit models},
  author={Song, Jiaming and Meng, Chenlin and Ermon, Stefano},
  booktitle={International Conference on Learning Representations},
  year={2021}
}

@article{jalal2021robust,
  title={Robust compressed sensing mri with deep generative priors},
  author={Jalal, Ajil and Arvinte, Marius and Daras, Giannis and Price, Eric and Dimakis, Alexandros G and Tamir, Jon},
  journal={Advances in neural information processing systems},
  volume={34},
  pages={14938--14954},
  year={2021}
}

@inproceedings{chung2022diffusion,
  title={Diffusion posterior sampling for general noisy inverse problems},
  author={Chung, Hyungjin and Kim, Jeongsol and Mccann, Michael T and Klasky, Marc L and Ye, Jong Chul},
  booktitle={International Conference on Learning Representations},
  year={2023}
}

@inproceedings{zhang2025improving,
  title={Improving diffusion inverse problem solving with decoupled noise annealing},
  author={Zhang, Bingliang and Chu, Wenda and Berner, Julius and Meng, Chenlin and Anandkumar, Anima and Song, Yang},
  booktitle={Proceedings of the Computer Vision and Pattern Recognition Conference},
  pages={20895--20905},
  year={2025}
}

@inproceedings{feng2023score,
  title={Score-based diffusion models as principled priors for inverse imaging},
  author={Feng, Berthy T and Smith, Jamie and Rubinstein, Michael and Chang, Huiwen and Bouman, Katherine L and Freeman, William T},
  booktitle={Proceedings of the IEEE/CVF International Conference on Computer Vision},
  pages={10520--10531},
  year={2023}
}

@inproceedings{cardoso2023monte,
  title={Monte Carlo guided diffusion for {B}ayesian linear inverse problems},
  author={Cardoso, Gabriel and Idrissi, Yazid Janati El and Corff, Sylvain Le and Moulines, Eric},
  booktitle={International Conference on Learning Representations},
  year={2024}
}

@article{li2024decoupled,
  title={Decoupled data consistency with diffusion purification for image restoration},
  author={Li, Xiang and Kwon, Soo Min and Liang, Shijun and Alkhouri, Ismail R and Ravishankar, Saiprasad and Qu, Qing},
  journal={arXiv preprint arXiv:2403.06054},
  year={2024}
}

@article{kawar2022DDRM,
  title={Denoising diffusion restoration models},
  author={Kawar, Bahjat and Elad, Michael and Ermon, Stefano and Song, Jiaming},
  journal={Advances in neural information processing systems},
  volume={35},
  pages={23593--23606},
  year={2022}
}

@inproceedings{
wang2023DDNM,
title={Zero-Shot Image Restoration Using Denoising Diffusion Null-Space Model},
author={Yinhuai Wang and Jiwen Yu and Jian Zhang},
booktitle={The Eleventh International Conference on Learning Representations },
year={2023},
}

@inproceedings{
song2024Resample,
title={Solving Inverse Problems with Latent Diffusion Models via Hard Data Consistency},
author={Bowen Song and Soo Min Kwon and Zecheng Zhang and Xinyu Hu and Qing Qu and Liyue Shen},
booktitle={The Twelfth International Conference on Learning Representations},
year={2024},
}

@article{xu2024PnP,
  title={Provably robust score-based diffusion posterior sampling for plug-and-play image reconstruction},
  author={Xu, Xingyu and Chi, Yuejie},
  journal={Advances in Neural Information Processing Systems},
  volume={37},
  pages={36148--36184},
  year={2024}
}

@InProceedings{Choi_2021_ICCV,
    author    = {Choi, Jooyoung and Kim, Sungwon and Jeong, Yonghyun and Gwon, Youngjune and Yoon, Sungroh},
    title     = {ILVR: Conditioning Method for Denoising Diffusion Probabilistic Models},
    booktitle = {Proceedings of the IEEE/CVF International Conference on Computer Vision (ICCV)},
    month     = {October},
    year      = {2021},
    pages     = {14367-14376}
}

@inproceedings{
xu2025rethinking,
title={Rethinking Diffusion Posterior Sampling: From Conditional Score Estimator to Maximizing a Posterior},
author={Tongda Xu and Xiyan Cai and Xinjie Zhang and Xingtong Ge and Dailan He and Ming Sun and Jingjing Liu and Ya-Qin Zhang and Jian Li and Yan Wang},
booktitle=ICLR,
year={2025}
}

@inproceedings{Yang2024Guidance,
author = {Yang, Lingxiao and Ding, Shutong and Cai, Yifan and Yu, Jingyi and Wang, Jingya and Shi, Ye},
title = {Guidance with spherical gaussian constraint for conditional diffusion},
year = {2024},
publisher = {JMLR.org},
booktitle = {Proceedings of the 41st International Conference on Machine Learning},
articleno = {2313},
numpages = {25},
location = {Vienna, Austria},
series = {ICML'24}
}

@article{chung2022improving,
  title={Improving diffusion models for inverse problems using manifold constraints},
  author={Chung, Hyungjin and Sim, Byeongsu and Ryu, Dohoon and Ye, Jong Chul},
  journal={Advances in Neural Information Processing Systems},
  volume={35},
  pages={25683--25696},
  year={2022}
}

@article{dhariwal2021diffusion,
  title={Diffusion models beat gans on image synthesis},
  author={Dhariwal, Prafulla and Nichol, Alexander},
  journal={Advances in neural information processing systems},
  volume={34},
  pages={8780--8794},
  year={2021}
}

@inproceedings{karras2019style,
  title={A style-based generator architecture for generative adversarial networks},
  author={Karras, Tero and Laine, Samuli and Aila, Timo},
  booktitle={Proceedings of the IEEE/CVF conference on computer vision and pattern recognition},
  pages={4401--4410},
  year={2019}
}

@inproceedings{deng2009imagenet,
  title={Imagenet: A large-scale hierarchical image database},
  author={Deng, Jia and Dong, Wei and Socher, Richard and Li, Li-Jia and Li, Kai and Fei-Fei, Li},
  booktitle={2009 IEEE conference on computer vision and pattern recognition},
  pages={248--255},
  year={2009},
  organization={Ieee}
}

@inproceedings{zhang2018unreasonable,
  title={The unreasonable effectiveness of deep features as a perceptual metric},
  author={Zhang, Richard and Isola, Phillip and Efros, Alexei A and Shechtman, Eli and Wang, Oliver},
  booktitle={Proceedings of the IEEE conference on computer vision and pattern recognition},
  pages={586--595},
  year={2018}
}

@article{heusel2017gans,
  title={Gans trained by a two time-scale update rule converge to a local nash equilibrium},
  author={Heusel, Martin and Ramsauer, Hubert and Unterthiner, Thomas and Nessler, Bernhard and Hochreiter, Sepp},
  journal={Advances in neural information processing systems},
  volume={30},
  year={2017}
}

@article{wang2004image,
  title={Image quality assessment: from error visibility to structural similarity},
  author={Wang, Zhou and Bovik, Alan C and Sheikh, Hamid R and Simoncelli, Eero P},
  journal={IEEE transactions on image processing},
  volume={13},
  number={4},
  pages={600--612},
  year={2004},
  publisher={IEEE}
}

@inproceedings{zhu2023denoising,
  title={Denoising diffusion models for plug-and-play image restoration},
  author={Zhu, Yuanzhi and Zhang, Kai and Liang, Jingyun and Cao, Jiezhang and Wen, Bihan and Timofte, Radu and Van Gool, Luc},
  booktitle={Proceedings of the IEEE/CVF conference on computer vision and pattern recognition},
  pages={1219--1229},
  year={2023}
}

@article{lu2022dpm,
  title={Dpm-solver: A fast ode solver for diffusion probabilistic model sampling in around 10 steps},
  author={Lu, Cheng and Zhou, Yuhao and Bao, Fan and Chen, Jianfei and Li, Chongxuan and Zhu, Jun},
  journal={Advances in neural information processing systems},
  volume={35},
  pages={5775--5787},
  year={2022}
}

@article{lu2025dpm,
  title={Dpm-solver++: Fast solver for guided sampling of diffusion probabilistic models},
  author={Lu, Cheng and Zhou, Yuhao and Bao, Fan and Chen, Jianfei and Li, Chongxuan and Zhu, Jun},
  journal={Machine Intelligence Research},
  pages={1--22},
  year={2025},
  publisher={Springer}
}

@book{tarantola2005inverse,
  title={Inverse problem theory and methods for model parameter estimation},
  author={Tarantola, Albert},
  year={2005},
  publisher={SIAM}
}

@inproceedings{shah2018solving,
  title={Solving linear inverse problems using gan priors: An algorithm with provable guarantees},
  author={Shah, Viraj and Hegde, Chinmay},
  booktitle={2018 IEEE international conference on acoustics, speech and signal processing (ICASSP)},
  pages={4609--4613},
  year={2018},
  organization={IEEE}
}

@article{AAP1238,
author = {Alain Durmus and {\'E}ric Moulines},
title = {{Nonasymptotic convergence analysis for the unadjusted Langevin algorithm}},
volume = {27},
journal = {The Annals of Applied Probability},
number = {3},
publisher = {Institute of Mathematical Statistics},
pages = {1551 -- 1587},
year = {2017}
}

@article{10.1111/rssb.12183,
    author = {Dalalyan, Arnak S.},
    title = {Theoretical Guarantees for Approximate Sampling from Smooth and Log-Concave Densities},
    journal = {Journal of the Royal Statistical Society Series B: Statistical Methodology},
    volume = {79},
    number = {3},
    pages = {651-676},
    year = {2016},
    month = {04},
    issn = {1369-7412}
}

@article{wu2024principled,
  title={Principled probabilistic imaging using diffusion models as plug-and-play priors},
  author={Wu, Zihui and Sun, Yu and Chen, Yifan and Zhang, Bingliang and Yue, Yisong and Bouman, Katherine L},
  journal={Advances in Neural Information Processing Systems},
  volume={37},
  pages={118389--118427},
  year={2024}
}

\clearpage
\setcounter{page}{1}
\setcounter{section}{1}
\setcounter{figure}{0}
\setcounter{table}{0}
\setcounter{equation}{0}
\renewcommand{\thesection}{S\arabic{section}}
\renewcommand{\thesubsection}{S\arabic{section}.\arabic{subsection}}
\renewcommand{\theequation}{S\arabic{equation}}
\renewcommand{\thefigure}{S\arabic{figure}}
\renewcommand{\thetable}{S\arabic{table}}

\begin{center}
    {\Large\bfseries Supplementary Material}
\end{center}
\vspace{1em}
\subsection{Lipschitz Analysis of Prior--Likelihood Interaction}
\label{sec:appendix:lipschitz}

To establish the negligible contribution of the prior gradient in high-noise regimes, we provide a Lipschitz-bound analysis quantifying the relative magnitudes of the likelihood and prior terms. We first introduce standard assumptions.

\paragraph{Assumptions.}
\begin{itemize}[noitemsep,topsep=2pt]
    \item[\textbf{A1.}] The score function $\nabla_{\mathbf{x}_t}\log p_t(\mathbf{x}_t;\sigma_t)$ is $L_{\sigma_t}$-Lipschitz continuous with $L_{\sigma_t} \le C/\sigma_t^{2}$ for some constant $C>0$, following standard smoothed density properties.
    \item[\textbf{A2.}] The measurement operator $\mathcal{A}$ is linear. We denote its smallest \textit{non-zero} singular value as $\sigma_{\min}^+(\mathcal{A}) > 0$.
    \item[\textbf{A3.}] The observation noise variance $\gamma^{2}$ is finite, with $\sigma_t \gg \gamma$ in high-noise regimes.
\end{itemize}

\paragraph{Prior negligibility via Lipschitz bounds.}
Consider the linear inverse problem
$\mathbf{y} = \mathcal{A}\mathbf{x}_{0} + \boldsymbol{\epsilon}$
where $\boldsymbol{\epsilon}\sim\mathcal{N}(0,\gamma^{2}\mathbf{I})$ and we define the measurement residual
\begin{equation}
\mathbf{r}_t := \mathbf{y} - \mathcal{A}\,\hat{\mathbf{x}}_0(\mathbf{x}_t).
\end{equation}
Under Assumptions~\textbf{A1}--\textbf{A3}, the gradient ratio satisfies
\begin{equation}
\kappa_t \;\ge\;
\frac{\sigma_{\min}^+(\mathcal{A})}{\gamma^{2} C}\,\sigma_t\,\|\mathbf{r}_t\|.
\label{eq:appendix:kappa_lower_bound}
\end{equation}
When $\sigma_t/\gamma \gg 1$ and $\|\mathbf{r}_t\|=\Theta(1)$, we obtain $\kappa_t\gg 1$, implying dominance of the likelihood gradient.

\paragraph{Proof.}
We upper-bound the prior gradient and lower-bound the likelihood gradient.

\textit{(i) Upper bound on the prior gradient.}
From Assumption~\textbf{A1}, the magnitude of the score is bounded by
\begin{equation}
\|\nabla_{\mathbf{x}_t}\log p_t(\mathbf{x}_t;\sigma_t)\|
\;\lesssim\; \frac{C}{\sigma_t}.
\label{eq:appendix:prior_upper}
\end{equation}

\textit{(ii) Lower bound on the likelihood gradient.}
Using the Gaussian likelihood and Tweedie's formula, the likelihood score is:
\begin{equation}
\nabla_{\mathbf{x}_t}\log p_t(\mathbf{y}\mid \mathbf{x}_t;\sigma_t)
= \frac{1}{\gamma^{2}}\mathcal{A}^{\top}\mathbf{r}_t + O(\sigma_t).
\end{equation}
Since $\mathbf{r}_t$ lies in the measurement space (the range of $\mathcal{A}$ plus noise), the back-projection via $\mathcal{A}^{\top}$ is bounded by the smallest non-zero singular value acting on the effective residual. Thus:
\begin{equation}
\|\nabla_{\mathbf{x}_t}\log p_t(\mathbf{y}\mid \mathbf{x}_t;\sigma_t)\|
\;\ge\; \frac{\sigma_{\min}^+(\mathcal{A})}{\gamma^{2}}\|\mathbf{r}_t\|.
\label{eq:appendix:likelihood_lower}
\end{equation}

\textit{(iii) Lower bound on gradient ratio.}
Combining~\eqref{eq:appendix:prior_upper} and~\eqref{eq:appendix:likelihood_lower},
\begin{equation}
\kappa_t
\;\gtrsim\;
\frac{\sigma_{\min}^+(\mathcal{A})}{\gamma^{2} C}\,
\sigma_t\,\|\mathbf{r}_t\|,
\end{equation}
establishing~\eqref{eq:appendix:kappa_lower_bound}. \hfill$\square$

\paragraph{Implications.} This analysis rigorously justifies the observation that the prior gradient contributes negligibly in high-noise regimes. For standard EDM schedulers in DAPS with $\sigma_{\max} \approx 100$ and low observation noise ($\gamma \approx 0.01$), the gradient ratio reaches magnitudes of $\kappa_t \approx 10^6$, rendering the prior term mathematically insignificant during early sampling. As illustrated in~\cref{fig:gradient_ratio}, this dominance persists throughout the diffusion process, where the likelihood gradient continues to outweigh the prior contribution even at the minimal noise levels ($\sigma_{\min} \approx 0.1$). Consequently, even for ill-posed problems where $\mathcal{A}$ has a null space, the likelihood gradient operates in the row space with such high magnitude (scaled by $1/\gamma^2$) that it dictates the update direction, validating the omission of the prior in the time-marginal step.

\subsection{MCMC Sampling with Warm Start}
\label{sec:warmstart}

\begin{figure}[t]
  \centering
  \includegraphics[width=0.55\linewidth]{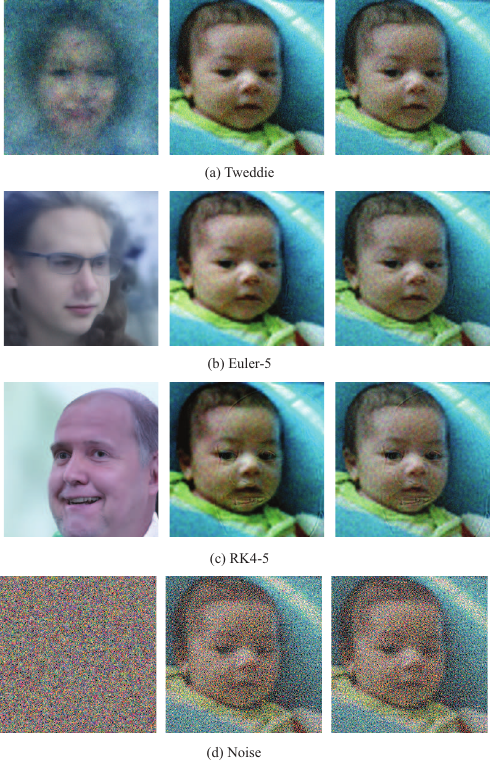}
  \caption{Visualization of 2-step MCMC iterations for $4\times$ downsampling under different initialization setups. Subfigures (a)--(c) are initialized at $\sigma_{\max} = 100$ using: (a) Tweedie's formula, (b) a 5-step Euler ODE solver, and (c) a 5-step RK-4 ODE solver. For comparison, (d) shows standard pure noise initialization at $\sigma= 1$.}
  \label{fig:mcmc_warmstart}
\end{figure}

When applying Markov chain Monte Carlo (MCMC), an informative initial state significantly improves both convergence speed and numerical accuracy. As shown in~\cref{fig:mcmc_warmstart}, different initialization strategies (including pure-noise initialization, high-order ODE-based initialization, and Tweedie-based estimation) are evaluated under identical MCMC settings (learning rate $1\times10^{-3}$, two refinement steps, and a $4\times$ downsampling operator). All schemes rapidly approach the measurement-consistent solution; however, the warm start obtained from Tweedie's estimator produces notably smoother updates with reduced oscillation. After the MCMC refinement step, the Tweedie-initialized trajectories exhibit markedly fewer high-frequency artifacts compared with those initialized via higher-order ODE solvers, demonstrating its advantage as a stable and noise-robust initialization method.

\subsection{Convergence Analysis}
\label{sec:convergence}

Building on the assumptions in~\cref{sec:appendix:lipschitz}, we show that the diffusion initialization provides a warm start of quality $\mathcal{O}(\sigma_{t})$ at every cycle, and that the ULA refinement converges in $\mathcal{O}(\kappa(\mathcal{A})^{2})$ steps.

\paragraph{Initialization quality.}
Both methods can be viewed as numerical integrators of the same probability flow ODE from $\sigma_{t}$ to $0$. Tweedie's formula corresponds to a single Euler step with step size $h=\sigma_{t}$ under a EDM diffusion~\cite{karras2022elucidating} setting, incurring a truncation error of $\mathcal{O}(\sigma_{t}^{2})$; an $N$-step RK4 solver with step size $h=\sigma_{t}/N$ reduces this to $\mathcal{O}(\sigma_{t}^{4}/N^{4})$. The threshold $\bar{\sigma}$ switches from the single-step Euler (Tweedie) to the higher-order solver when the first-order truncation error begins to exceed the measurement noise floor~$\gamma$.

\paragraph{Warm-start convergence of ULA.}
Under Assumption \textbf{A2}, the negative log likelihood satisfies $\gamma^{-2}\mathcal{A}^{\top}\mathcal{A}$, which is $\mu$-strongly convex and $L$-smooth when restricted to $\mathrm{Row}(\mathcal{A})$, with
\begin{equation}
    \mu = \frac{[\sigma_{\min}^{+}(\mathcal{A})]^{2}}{\gamma^{2}},
    \qquad
    L = \frac{\|\mathcal{A}\|^{2}}{\gamma^{2}}.
\end{equation}
Starting from $\hat{\mathbf{x}}_0$ at distance $e_{t_k}$ from measurement-consistent solution $\mathbf{x}^{*}$, the step size $\eta\le 1/L$ satisfies the standard $W_2$ convergence bound~\cite{10.1111/rssb.12183,AAP1238}:
\begin{equation}
    W_2\!\bigl(\mathrm{Law}(x^{(K)}),\,\pi\bigr)
    \;\le\;
    (1-\eta\mu)^{K}\,e_{t_k}
    \;+\;
    \mathcal{O}\!\!\left(\sqrt{\frac{\eta\,d_{\mathrm{row}}}{\mu}}\right),
    \label{eq:ula_w2}
\end{equation}
where $\pi\propto\exp(-\ell)$ is the target likelihood distribution and $d_{\mathrm{row}}=\mathrm{rank}(\mathcal{A})$. The first term captures exponential \emph{contraction} of the initial error; the second is the \emph{discretization bias} inherent to ULA.

To determine the required step count, we set the contraction term below a target tolerance~$\delta$:
\begin{equation}
    (1-\eta\mu)^{K}\,e_{t_k} \;\le\; \delta
    \quad\Longrightarrow\quad
    K\,\ln(1-\eta\mu) \;\le\; \ln\frac{\delta}{e_{t_k}}.
\end{equation}
Applying $\ln(1-x)\le -x$ for $x=\eta\mu\in(0,1)$ and rearranging gives 
\begin{equation}
K \ge \frac{1}{\eta\mu}\ln\frac{e_{t_k}}{\delta}.
\end{equation}
Substituting the step size $\eta=1/L$ and noting that $L/\mu = \|\mathcal{A}\|^{2}/[\sigma_{\min}^{+}(\mathcal{A})]^{2} = \kappa(\mathcal{A})^{2}$:
\begin{equation}
    K
    \;=\;
    \mathcal{O}\!\left(\frac{1}{\eta\mu}\,\log\frac{e_{t_k}}{\delta}\right)
    \;=\;
    \mathcal{O}\!\left(\kappa(\mathcal{A})^{2}\,\log\frac{e_{t_k}}{\delta}\right),
    \label{eq:K_general}
\end{equation}

In Algorithm~1, re-noising at the end of each cycle adds noise of magnitude~$\sigma_{t_{k+1}}$, so corrections below this scale are erased. Setting $\delta=\mathcal{O}(\sigma_{t_{k+1}})$ and substituting $e_{t_k}=\mathcal{O}(\sigma_{t_k})$:
\begin{equation}
    K
    \;=\;
    \mathcal{O}\!\left(
        \kappa(\mathcal{A})^{2}\,\log\frac{\sigma_{t_k}}{\sigma_{t_{k+1}}}
    \right)
    \;=\;
    \mathcal{O}\!\bigl(\kappa(\mathcal{A})^{2}\,\log\rho\bigr),
    \label{eq:K_percycle}
\end{equation}
where $\rho=\sigma_{t_k}/\sigma_{t_{k+1}}$ is the annealing ratio. For geometric schedules, $\rho$ is constant, and the dominant computational factor is $\kappa(\mathcal{A})^{2}$, which is determined by the spectral properties of the forward operator.

\paragraph{Convergence across problem types.}
For nonlinear operators, $\kappa(\mathcal{A})$ in~\eqref{eq:K_percycle} is replaced by the local effective condition number $\kappa_{\mathrm{eff}}=\|\nabla\mathcal{A}(\hat{\mathbf{x}}_0)\|/\sigma_{\min}^{+}(\nabla\mathcal{A}(\hat{\mathbf{x}}_0))$ of the Jacobian at the initialization. This explains the empirical variation in refinement steps: well-conditioned operators such as downsampling ($\kappa_{\mathrm{eff}}$ small, $K=2$) converge rapidly, while operators with broader eigenvalue spread such as Gaussian deblurring require more steps. For strongly nonlinear operators where the local likelihood is non-convex (e.g., phase retrieval), the strong-convexity assumption breaks down and local refinement may converge to an incorrect basin.

\section{Discussions}
\subsection{Sampling Efficiency}
\begin{figure}[t]
    \centering
    \includegraphics[width=\linewidth]{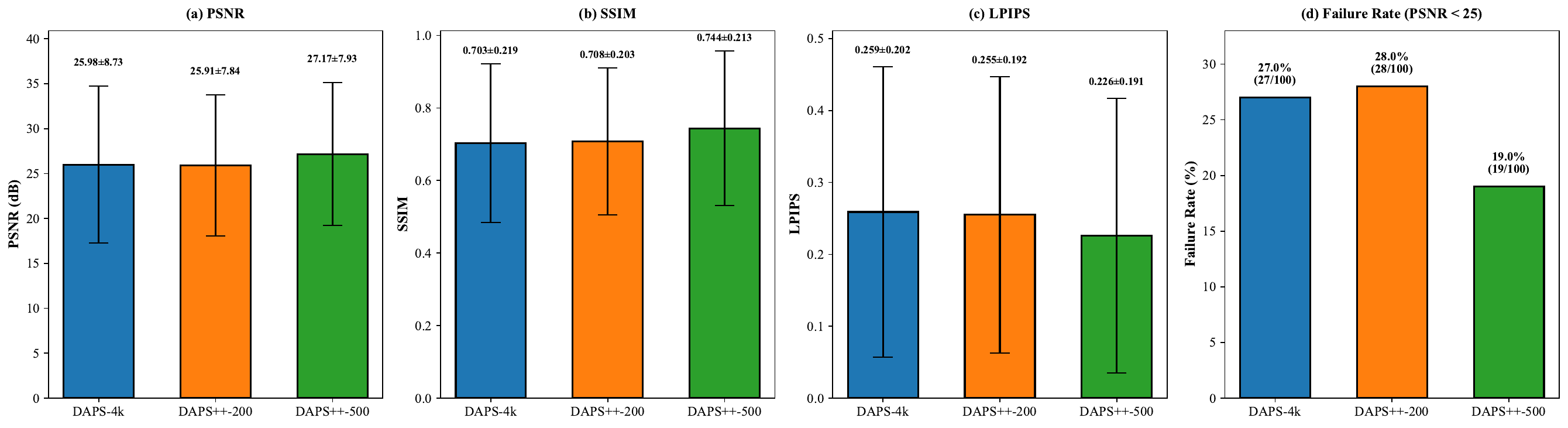}
    \caption{Phase retrieval results using the top 100 images from the FFHQ test dataset.
    (a)~PSNR, (b)~SSIM, and (c)~LPIPS statistics (mean $\pm$ standard deviation) show that DAPS++ achieves reconstruction quality comparable to DAPS-4K while requiring substantially fewer sampling steps.
    (d)~Failure rate (PSNR $<$ 25) shows that instability arises from the nonlinear frequency-domain measurement operator.}
    \label{fig:phase_variance}
\end{figure}
Sampling efficiency is a critical factor for diffusion-based inverse problem solvers.
The computational cost of these methods depends heavily on both the number of neural
function evaluations (NFEs) and the refinement steps performed at each iteration.
In~\cref{tab:SR_time}, we summarize the ODE steps, annealing steps, refinement steps,
and resulting NFE budgets for several pixel-space baseline methods and for DAPS under
multiple configurations, along with the corresponding single-image sampling times on
FFHQ-256, compared against DAPS++. As shown, DAPS++ achieves substantially higher
efficiency than existing baselines, delivering competitive performance even with very
low NFE budgets.
\begin{table}[t]
  \caption{Sampling time of DAPS on the super-resolution task with FFHQ-256 based on $\bar{\sigma} = 0.5$.
  Reported values denote non-parallel single-image sampling time on the FFHQ-256 dataset using one NVIDIA A100 PCIe 80GB GPU.}
  \label{tab:SR_time}
  \centering
  \small
  \setlength{\tabcolsep}{4pt}
  \begin{tabular}{lccccc}
    \toprule
    Config. & ODE Steps & Anneal. Steps & Refine. Steps & NFE & Sec./Img. \\
    \midrule
    DPS      & -- & --  & 1   & 1000 & 40.90 \\
    DDRM     & -- & --  & --  & 20   & 0.58  \\
    DiffPIR  & -- & --  & --  & 100  & 2.41  \\
    \midrule
    DAPS-100    & 2 & 50  & 100 & 100  & 6.26  \\
    DAPS-1K     & 5 & 200 & 100 & 1000 & 38.10 \\
    DAPS++-20   & 1 & 20  & 2   & 40   & 0.89  \\
    DAPS++-50   & 1 & 50  & 2   & 100  & 2.31  \\
    DAPS++-100  & 1 & 100 & 2   & 200  & 4.62  \\
    DAPS++-200  & 1 & 200 & 2   & 400  & 9.25  \\
    \bottomrule
  \end{tabular}
\end{table}
\subsection{Regarding Phase Retrieval}
\label{sec:pr}
\begin{figure}[t]
  \centering
  \includegraphics[width=0.75\linewidth]{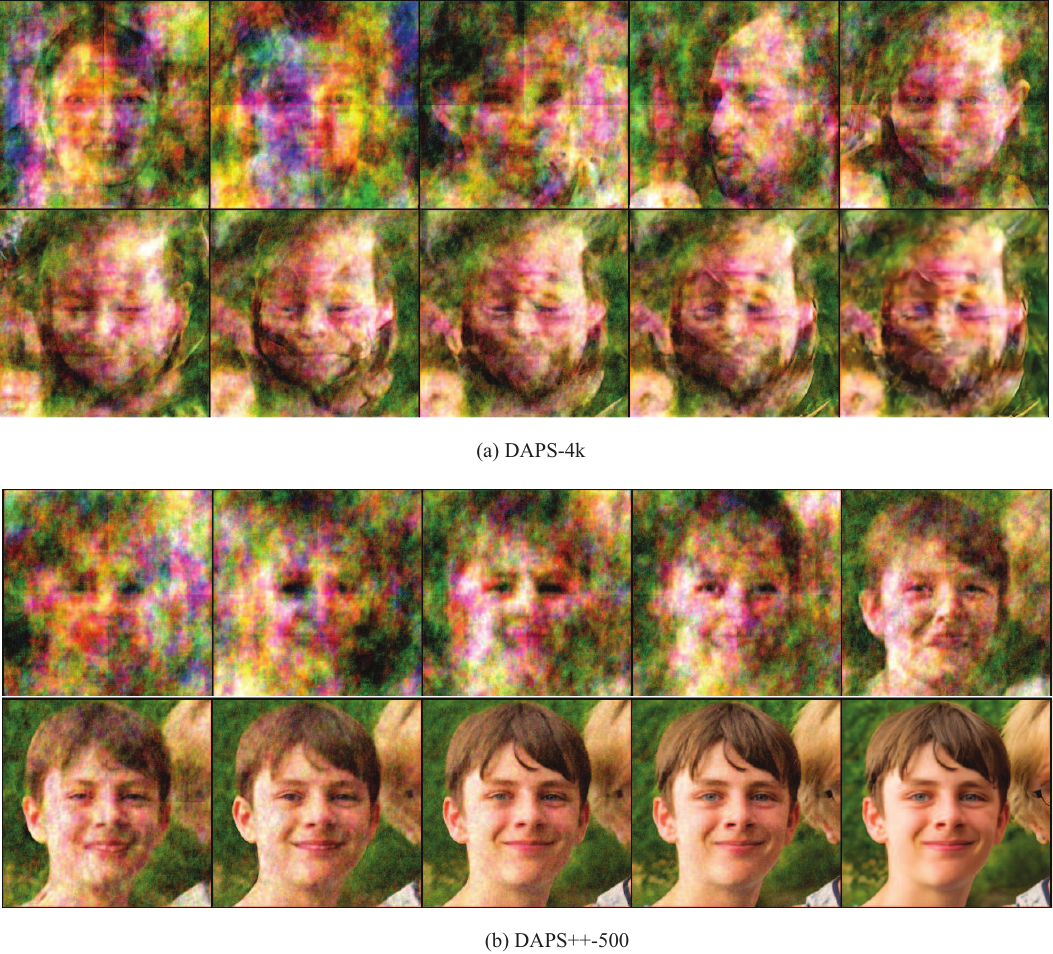}
  \caption{Visualization of the trajectory of diffusion outputs in the phase retrieval task for
(a)~DAPS-4K and (b)~DAPS++-500.}
  \label{fig:phase_variance_tj}
\end{figure}

DAPS has demonstrated strong performance on the phase retrieval task compared with other baselines. However, this advantage is not consistently maintained across the entire dataset. As illustrated in~\cref{fig:phase_variance}, the variance of failure cases among the top 100 images from the FFHQ test set remains significant. These failures primarily stem from the nonlinear frequency-domain measurement operator, which interacts unfavorably with the pixel-domain initialization produced by the diffusion prior. A trajectory comparison of the diffusion output $\hat{x}_0$ in~\cref{fig:phase_variance_tj} further shows that DAPS++ exhibits more stable refinement behavior and achieves more reliable reconstructions under the same experimental conditions.
Despite this challenge, DAPS++ achieves performance comparable to DAPS-4K while
using substantially fewer sampling steps, and it exhibits a noticeably lower
failure rate. This highlights that DAPS++ does not introduce mechanisms that address frequency-domain nonlinearities, and its behavior remains constrained in the same way as DAPS. However, with more annealing steps, the initialization has a higher chance of producing a sample closer to the true solution under a longer noise schedule.

\subsection{Future Directions}
\label{sec:future_directions}

By decoupling the solution into two iterative and independent stages, our framework avoids the complex requirement of treating the likelihood term $p(\mathbf{y} \mid \hat{\mathbf{x}}_0(\mathbf{x}_t))$ as an intrinsic part of the diffusion-induced probability manifold.
Instead, each stage can be analyzed and optimized in isolation.
Specifically, the refinement phase is liberated from the constraints of maintaining time-marginal consistency or strictly adhering to the diffusion trajectory.
This independence provides a more flexible and theoretically robust foundation for applying diffusion models to measurement-driven reconstruction.

Consequently, this formulation naturally extends to complex, non-Gaussian noise models—such as Poisson or Speckle noise—which have historically been difficult to incorporate into coupled diffusion updates.
Since the diffusion model serves primarily to provide initialization and regularize the feasible solution space, such noise distributions become mathematically tractable.
Future work will explore leveraging this decoupled architecture to handle these diverse noise models in a principled and elegant manner.

\section{Experimental Details}
We follow the forward measurement operators utilized in DAPS~\cite{zhang2025improving} and Resample~\cite{song2024Resample}, establishing a unified evaluation protocol for general linear and nonlinear inverse problems.

\subsection{Final Refinement via RK4 Solver}
\label{sec:rk4_refinement}

To achieve high-fidelity reconstruction, we introduce a final refinement stage subsequent to the initial sampling process (which operates in the regime $\sigma_t \le \bar{\sigma}$).
Specifically, we adopt the formulation from Karras \etal~\cite{karras2022elucidating} (EDM) with $\alpha_t \equiv 1$ (Variance Exploding SDE) and perform deterministic refinement by integrating the probability-flow ODE.
The drift term $\mathbf{f}(\mathbf{x}_t,\sigma_t)$ governing this trajectory is defined as
\begin{equation}
    \mathbf{f}(\mathbf{x}_t,\sigma_t)
    = -\dot{\sigma}_t\,\sigma_t\, s_\theta(\mathbf{x}_t,\sigma_t),
\end{equation}
where $s_\theta(\mathbf{x}_t,\sigma_t) \approx \nabla_{\mathbf{x}_t}\log p_t(\mathbf{x}_t)$ denotes the diffusion model's score estimate.
We employ a fourth-order Runge--Kutta (RK4) solver to integrate this ODE specifically for these final refinement steps.
Given the current state $\mathbf{x}_t$ and noise level $\sigma_t$, a single RK4 update with step size $h$ is computed as
\begin{align}
    k_1 &= \mathbf{f}(\mathbf{x}_t,\sigma_t), \notag\\
    k_2 &= \mathbf{f}\!\left(\mathbf{x}_t + \tfrac{h}{2}k_1,\ \sigma_t + \tfrac{h}{2}\dot{\sigma}_t\right), \notag\\
    k_3 &= \mathbf{f}\!\left(\mathbf{x}_t + \tfrac{h}{2}k_2,\ \sigma_t + \tfrac{h}{2}\dot{\sigma}_t\right), \notag\\
    k_4 &= \mathbf{f}\!\left(\mathbf{x}_t + hk_3,\ \sigma_t + h\dot{\sigma}_t\right), \notag\\
    \mathbf{x}_{t-h} &= \mathbf{x}_t + \frac{h}{6}\!\left(k_1 + 2k_2 + 2k_3 + k_4\right).
    \label{eq:rk4}
\end{align}
\begin{table}[t]
\caption{\textbf{Hyperparameter comparison between DAPS and DAPS++.}
We list the step size $\eta_0$, noise scale $\delta$, and total sampling steps. Note that while DAPS requires 100 steps for all tasks, DAPS++ achieves efficient reconstruction with significantly fewer steps (\eg, 2 steps for Super-Resolution).}
\label{tab:hyperparameters}
\centering
\resizebox{0.95\textwidth}{!}{%
\begin{tabular}{l|c|cccccc}
\toprule
\multicolumn{1}{l}{\textbf{Algorithm}} &
\multicolumn{1}{c}{\textbf{Params}} &
\textbf{SR 4$\times$} &
\textbf{Inpainting} &
\textbf{Gauss. Deblur} &
\textbf{Motion Deblur} &
\textbf{Nonlinear Deblur} &
\textbf{HDR} \\
\midrule

\textbf{DAPS} & $\eta_0$ & 1e-4 & 1e-4 & 1e-4 & 5e-5 & 5e-5 & 2e-5 \\
              & $\delta$ & 1e-2 & 1e-2 & 1e-2 & 1e-2 & 1e-2 & 1e-2 \\
              & Steps    & 100  & 100  & 100  & 100  & 100  & 100  \\
\midrule

\textbf{DAPS++ (Ours)} & $\eta_0$ & 1e-3 & 1e-4 & 1e-4 & 1e-4 & 1e-5  & 2.5e-5 \\
                       & $\delta$ & 1e-2 & 1e-2 & 1e-2 & 1e-2 & 1e-2 & 1e-2  \\
                       & \textbf{Steps} & \textbf{2} & \textbf{5} & \textbf{8} & \textbf{8} & \textbf{50} & \textbf{5} \\
\bottomrule
\end{tabular}}
\end{table}

\begin{figure}[t]
  \centering
  \includegraphics[width=0.7\linewidth]{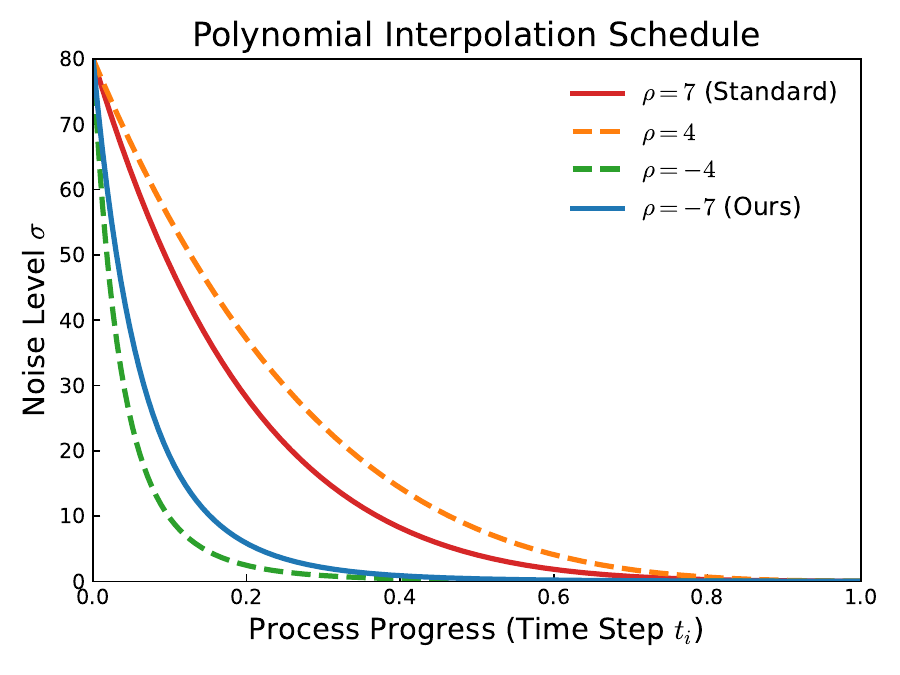}
\caption{\textbf{Polynomial interpolation schedules.} The standard $\rho=7$ (red) maintains high noise levels, whereas our $\rho=-7$ (blue) accelerates decay. This allocates more steps to the small-noise regime, crucial for detailed refinement in inverse problems.}
  \label{fig:schedule}
\end{figure}

\noindent
To discretize the time variable during this refinement phase, we employ the polynomial interpolation schedule from EDM~\cite{karras2022elucidating}.
For the $i$-th step among $N$ total steps, the discretized time $t_i$ is determined by
\begin{equation}
    t_i
    = t_{\min}
    + \bigl(t_1 - t_{\min}\bigr)
      \left(\frac{i}{N-1}\right)^{\rho}.
    \label{eq:poly-schedule}
\end{equation}
While standard diffusion training typically utilizes $\rho = 7$ and $t_{\min} = 0.02$, we tailor the schedule for the inference phase of DAPS++ (specifically with $\gamma=0.05$) to better suit the refinement requirements in the inverse problem space.
Accordingly, we set $\rho = -7$ to induce a faster decay of the noise level, thereby concentrating a greater density of discretization steps within the small-noise regime, as illustrated in~\cref{fig:schedule}.
With this schedule, given the refinement threshold $\bar{\sigma}=0.5$, approximately 30\% of the total sampling steps are allocated to the refinement stage ($\sigma_t \le \bar{\sigma}$), ensuring detailed reconstruction.

\subsection{Hyperparameter Choice}
\label{sec:hyperparameters}

To evaluate the proposed method across general linear and nonlinear inverse problems, we maintain a unified hyperparameter setting to demonstrate robustness.
Detailed configurations for all tasks, which include Super-Resolution ($4\times$), Inpainting, Deblurring (Gaussian/Motion/Nonlinear), and High Dynamic Range (HDR), are summarized in~\cref{tab:hyperparameters}, where we provide a direct comparison with the baseline DAPS~\cite{zhang2025improving}.

\noindent\textbf{Efficiency and NFEs.}
For the annealing sampling process, we strictly limit computation to a single-step solver per time step.
Unlike methods that require multiple gradient steps or extensive MCMC transitions at each noise level, our approach achieves fast and robust reconstruction with minimal computational overhead.
As shown in~\cref{tab:hyperparameters}, while the original DAPS requires a fixed budget of $100$ MCMC steps for all tasks, DAPS++ significantly reduces the inference cost without compromising quality.

\noindent\textbf{Annealing Steps for Different Datasets.}
As reported in~\cref{tab:imagenet,tab:ffhq}, we adjust the total number of annealing steps ($K$) based on the dataset complexity.
Specifically, we employ a higher number of steps for ImageNet compared to FFHQ.
This increase is necessitated by the significantly higher semantic diversity and manifold complexity of ImageNet, which requires a finer discretization of the reverse trajectory to prevent the sampling path from diverging.
Conversely, FFHQ, being a domain-specific face dataset with lower entropy, allows for faster convergence with fewer steps.

\noindent\textbf{General Parameters and Decay Schedule.}
The step size $\eta_0$ and noise level $\delta$ are tuned per task to balance data consistency and perceptual quality.
For the step size $\eta_t$ at timestep $t$, we adopt a linear decay scheme defined as:
\begin{equation}
    \eta_t = \eta_0 \left[ \delta + \frac{t}{T}(1 - \delta) \right],
\end{equation}
where $\delta$ represents the decay ratio and $T$ is the starting timestep.
Furthermore, regarding the measurement noise parameter $\gamma$, we treat it as an effective noise level rather than strictly adhering to the physical ground truth.
Regardless of the true noise level variation (\eg, ranging from $\gamma=0.05$ to $0.4$), we consistently set the hyperparameter to $\gamma = 0.01$ to achieve better empirical performance and stability during the reconstruction process.

\section{More Ablation Results}
\label{sec:more_ablation}

\begin{table}[t]
\caption{Quantitative comparison for varying $\bar{\sigma}$ on 100 validation images of \textbf{FFHQ}.
Metrics are reported as \textbf{SSIM$\uparrow$ / LPIPS$\downarrow$ / FID$\downarrow$}.
\textbf{Bold} indicates the best result; \underline{underlined} values denote the second best.
All measurements include additive Gaussian noise with $\gamma = 0.05$ and $\rho = -7$.}
\label{tab:sigma_comparison}
 \centering
 \renewcommand{\arraystretch}{1.3}
 \resizebox{\linewidth}{!}{
 \begin{tabular}{lccccccc}
 \toprule
 \textbf{$\bar{\sigma}$} &
 \textbf{SR $\times$4} &
 \textbf{Inpainting} &
 \textbf{Gaussian Blur} &
 \textbf{Motion Blur} &
 \textbf{Nonlinear Blur} &
 \textbf{HDR} &
 \textbf{NFE} \\
 \cmidrule(lr){2-7}
  & SSIM$\uparrow$ / LPIPS$\downarrow$ / FID$\downarrow$ &
    SSIM$\uparrow$ / LPIPS$\downarrow$ / FID$\downarrow$ &
    SSIM$\uparrow$ / LPIPS$\downarrow$ / FID$\downarrow$ &
    SSIM$\uparrow$ / LPIPS$\downarrow$ / FID$\downarrow$ &
    SSIM$\uparrow$ / LPIPS$\downarrow$ / FID$\downarrow$ &
    SSIM$\uparrow$ / LPIPS$\downarrow$ / FID$\downarrow$ &
    \\
 \midrule
 0.2 &
 \textbf{0.798} / 0.182 / 52.2 &
 \textbf{0.819} / \textbf{0.138} / 43.3 &
 \textbf{0.799} / 0.185 / 60.1 &
 \textbf{0.836} / 0.144 / 43.6 &
 \textbf{0.750} / 0.204 / 63.0 &
 \textbf{0.832} / 0.174 / 45.7 &
74 \\

 0.5 &
 \underline{0.782} / \underline{0.176} / 48.6 &
 \underline{0.814} / \textbf{0.138} / \underline{41.9} &
 \underline{0.783} / 0.171 / 51.2 &
 \underline{0.828} / \underline{0.136} / 38.7 &
 0.744 / 0.195 / 52.6 &
 \underline{0.831} / 0.171 / 42.8 &
100 \\

 1.0 &
 0.774 / \textbf{0.175} / \underline{46.7} &
 0.811 / \underline{0.139} / \textbf{41.8} &
 0.774 / \underline{0.168} / \underline{48.7} &
 0.826 / \textbf{0.135} / \textbf{37.9} &
 0.744 / 0.193 / \underline{51.4} &
 0.831 / \underline{0.169} / \underline{41.4} &
116 \\

 2.0 &
 0.772 / 0.175 / \textbf{45.6} &
 0.808 / 0.140 / 42.6 &
 0.771 / \textbf{0.167} / \textbf{47.7} &
 0.826 / 0.135 / 38.0 &
 0.745 / \underline{0.192} / 51.5 &
 0.829 / \textbf{0.168} / \textbf{40.5} &
134 \\

 5.0 &
 0.771 / 0.176 / 45.6 &
 0.803 / 0.143 / 42.9 &
 0.770 / 0.168 / 47.7 &
 0.826 / 0.135 / 37.9 &
 \underline{0.746} / \textbf{0.191} / \textbf{51.3} &
 0.790 / 0.210 / 55.7 &
152 \\
 \bottomrule
 \end{tabular}}
\end{table}

\begin{figure}[t]
  \centering
  \includegraphics[width=1\linewidth]{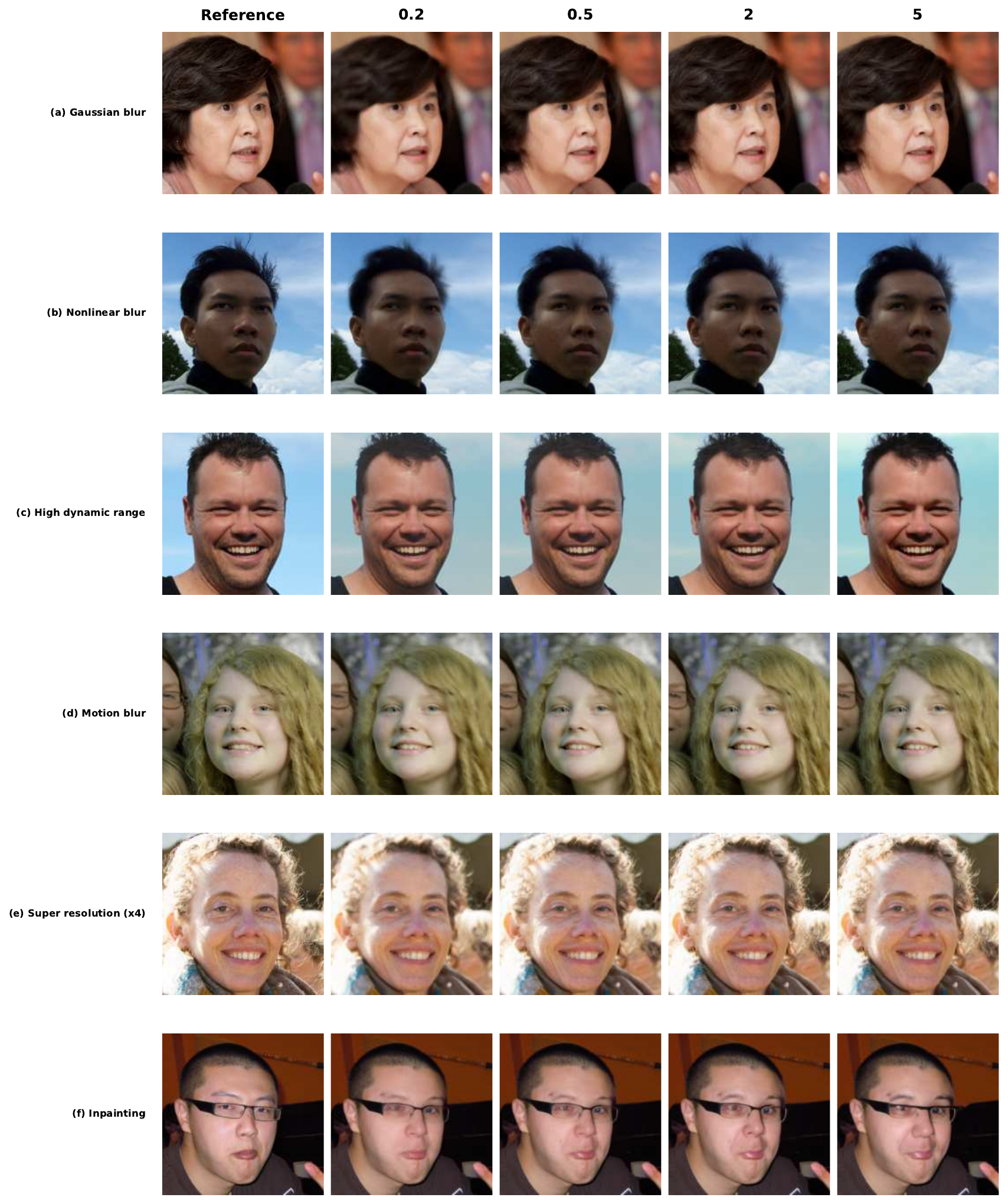}
  \caption{Qualitative results on representative inverse problems on the FFHQ-256 dataset. We compare reconstructions across varying noise refinement threshold $\bar{\sigma} \in \{0.2, 0.5, 2.0, 5.0\}$. The tasks are: (a)~Gaussian blur reconstruction; (b)~Nonlinear blur reconstruction; (c)~High dynamic range (HDR) reconstruction; (d)~Motion blur reconstruction; (e)~$4\times$ super-resolution; and (f)~Box inpainting.}
  \label{fig:sigma_comparison}
\end{figure}

\begin{table}[t]
\caption{Quantitative comparison for varying $\bar{\rho}$ on 100 validation images of \textbf{FFHQ}.
Metrics are reported as \textbf{SSIM$\uparrow$ / LPIPS$\downarrow$ / FID$\downarrow$}.
\textbf{Bold} indicates the best result; \underline{underlined} values denote the second best.
All measurements include additive Gaussian noise with $\gamma = 0.05$ and $\bar{\sigma} = 0.5$.}
\label{tab:rho_comparison}
 \centering
 \renewcommand{\arraystretch}{1.3}
 \resizebox{\linewidth}{!}{
 \begin{tabular}{lccccccc}
 \toprule
 \textbf{$\rho$} &
 \textbf{SR ×4} &
 \textbf{Inpainting} &
 \textbf{Gaussian Blur} &
 \textbf{Motion Blur} &
 \textbf{Nonlinear Blur} &
 \textbf{HDR} &
 \textbf{NFE} \\
 \cmidrule(lr){2-7}
  & SSIM$\uparrow$ / LPIPS$\downarrow$ / FID$\downarrow$ &
    SSIM$\uparrow$ / LPIPS$\downarrow$ / FID$\downarrow$ &
    SSIM$\uparrow$ / LPIPS$\downarrow$ / FID$\downarrow$ &
    SSIM$\uparrow$ / LPIPS$\downarrow$ / FID$\downarrow$ &
    SSIM$\uparrow$ / LPIPS$\downarrow$ / FID$\downarrow$ &
    SSIM$\uparrow$ / LPIPS$\downarrow$ / FID$\downarrow$ &
    \\
 \midrule
-2 &
 0.772 / 0.192 / 56.4 &
 \underline{0.811} / \textbf{0.141} / 44.4 &
 \textbf{0.790} / \textbf{0.162} / \textbf{46.3} &
 \textbf{0.840} / \textbf{0.126} / \textbf{32.5} &
 \textbf{0.751} / \textbf{0.187} / \textbf{51.2} &
 \underline{0.833} / \underline{0.174} / 45.9 &
134 \\

 -5 &
 \underline{0.780} / \underline{0.177} / \underline{47.2} &
 \textbf{0.812} / 0.141 / \underline{42.7} &
 \underline{0.785} / \underline{0.169} / \underline{50.4} &
 \underline{0.831} / \underline{0.134} / \underline{36.5} &
 \underline{0.747} / \underline{0.192} / \underline{51.4} &
 \textbf{0.834} / \textbf{0.169} / \textbf{42.0} &
107 \\

 -7 &
 \textbf{0.781} / \textbf{0.176} / \textbf{46.0} &
 0.812 / 0.141 / \textbf{42.1} &
 0.784 / 0.171 / 51.1 &
 0.829 / 0.136 / 37.9 &
 0.744 / 0.195 / 52.6 &
 0.834 / 0.169 / \underline{42.3} &
 100 \\

 2 &
 0.708 / 0.261 / 80.1 &
 0.793 / 0.170 / 50.5 &
 0.729 / 0.233 / 71.9 &
 0.770 / 0.198 / 60.2 &
 0.680 / 0.275 / 78.6 &
 0.790 / 0.213 / 52.5 &
 56 \\

 5 &
 0.772 / 0.196 / 55.0 &
 0.810 / 0.147 / 44.0 &
 0.767 / 0.195 / 58.9 &
 0.807 / 0.159 / 47.9 &
 0.720 / 0.223 / 63.9 &
 0.820 / 0.184 / 46.5 &
71 \\

 7 &
 0.778 / 0.189 / 52.2 &
 0.811 / \underline{0.145} / 44.4 &
 0.772 / 0.190 / 57.4 &
 0.812 / 0.154 / 45.9 &
 0.727 / 0.215 / 62.3 &
 0.823 / 0.182 / 46.2 &
74 \\

 \bottomrule
 \end{tabular}}
\end{table}

\begin{figure}[t]
  \centering
  \includegraphics[width=\linewidth]{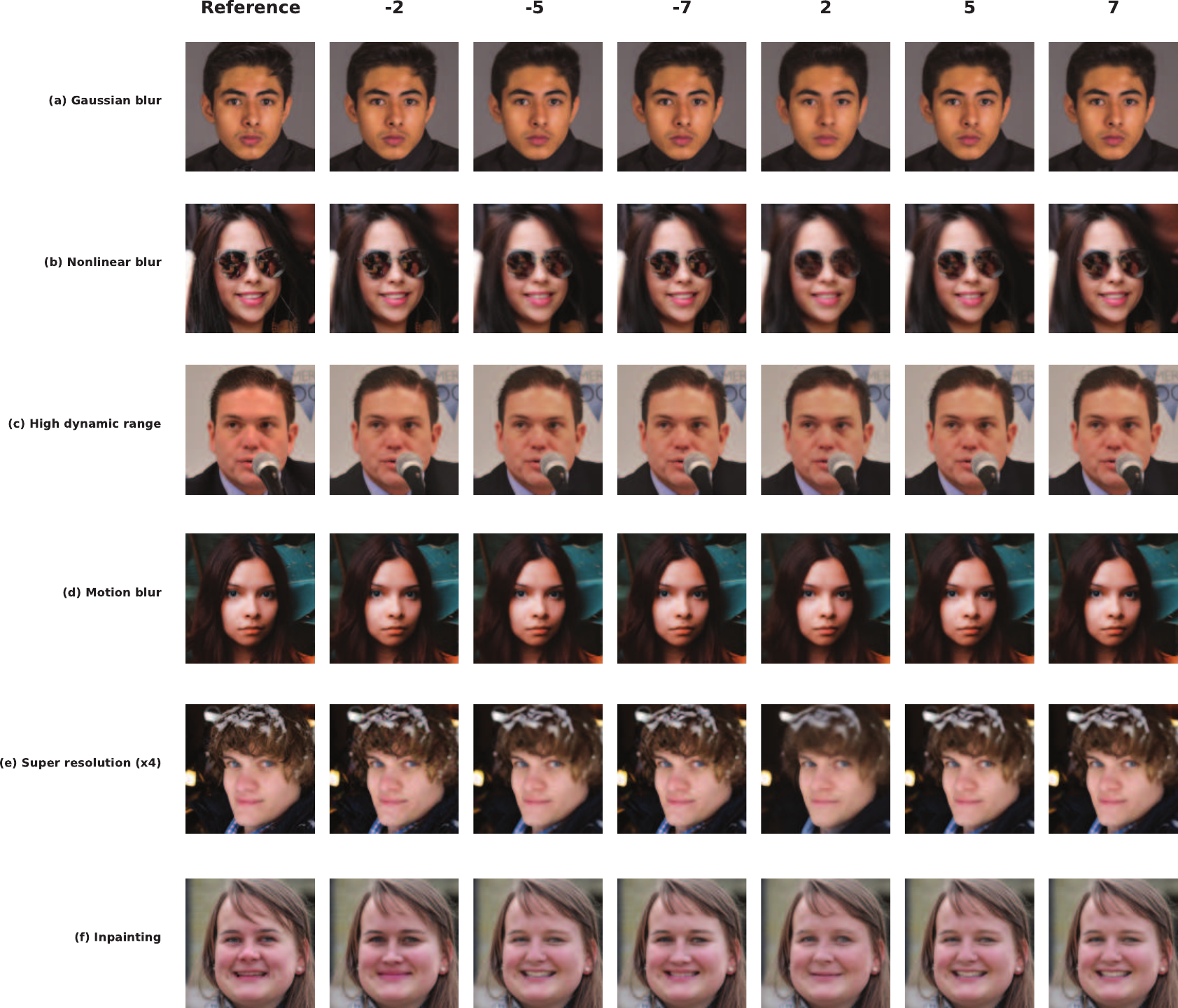}
  \caption{Qualitative results on representative inverse problems on the FFHQ-256 dataset. We compare reconstructions across varying polynomial interpolation schedules $\rho \in \{-2, -5, -7, 2,5,7\}$. The tasks are: (a)~Gaussian blur reconstruction; (b)~Nonlinear blur reconstruction; (c)~High dynamic range (HDR) reconstruction; (d)~Motion blur reconstruction; (e)~$4\times$ super-resolution; and (f)~Box inpainting.}
  \label{fig:rho_comparison}
\end{figure}

We present additional ablation studies on the choice of $\bar{\sigma}$ and the polynomial exponent $\rho$ in the noise schedule. Results across several inverse problems are summarized in~\cref{tab:sigma_comparison,tab:rho_comparison}, evaluated under identical measurement noise levels. Illustrative examples are provided in~\cref{fig:sigma_comparison,fig:rho_comparison}.

As shown in~\cref{tab:sigma_comparison}, using a larger $\bar{\sigma}$ increases the number of required NFEs but does not yield noticeable improvements in reconstruction quality; in some tasks, it even slightly degrades performance due to excessive early-stage denoising. Conversely, a smaller $\bar{\sigma}$ tends to improve SSIM but may oversuppress fine structures, reducing perceptual fidelity.

For the polynomial schedule,~\cref{tab:rho_comparison} shows that negative values of $\rho$ consistently enhance performance across tasks, indicating that spending more iterations in the low-noise regime benefits the refinement step. This behavior aligns with our decoupled initialization-refinement perspective: once the diffusion model provides a reliable initialization, allocating more resolution to small-noise updates improves measurement consistency without sacrificing stability.

\end{document}